\documentclass[runningheads]{llncs}

\AtBeginDocument{%
  }

\usepackage{overpic}
\usepackage{amsmath}
\usepackage{todonotes}
\usepackage{mathrsfs}
\usepackage{booktabs}
\usepackage{subcaption}
\usepackage{multicol}
\usepackage{tikz}
\usetikzlibrary{calc}
\usetikzlibrary{decorations.pathmorphing}

\usepackage{multirow}
\usepackage{amssymb}
\usepackage{hyperref}
\usepackage[commandnameprefix=always]{changes}
\usepackage{xcolor}

\newcommand{\ttrain}{\mathcal{T}_\text{train}}
\newcommand{\tval}{\mathcal{T}_\text{val}}
\newcommand{\ttest}{\mathcal{T}_\text{test}}

\begin{document}

\title{Oracle, will I ever learn? A study of prediction convergence and complementarity across link prediction models}
\titlerunning{Oracle, will I ever learn?}
% If the paper title is too long for the running head, you can set
% an abbreviated paper title here
%
\author{Guillaume Méroué\orcidID{0009-0004-7111-3202} \and
Fabien Gandon\orcidID{0000-0003-0543-1232} \and
Pierre Monnin\orcidID{0000-0002-2017-8426}}
\authorrunning{Méroué et al.}
% First names are abbreviated in the running head.
% If there are more than two authors, 'et al.' is used.
%
\institute{Université Côte d’Azur, Inria, CNRS, I3S, France \\
\email{firstname.lastname@inria.fr}}
\maketitle              % typeset the header of the contribution

\begin{abstract}

Knowledge graphs have become an important source of structured knowledge for Web applications, including search, question answering, and recommender systems.
In these applications, link prediction can serve either as a prediction task itself or as a means to enrich incomplete knowledge graphs for downstream tasks.
Interestingly, different link prediction models, or even different training runs of the same model, can produce substantially different predictions for the same query. This suggests a variability in the capture of the underlying knowledge by models, thus raising a fundamental question: to what extent do different models capture complementary knowledge, and how much of this knowledge could be recovered by combining them?
We propose to measure model complementarity through the performance of an oracle that, for each query, selects the best prediction among a considered set of models, hence providing an upper bound on the performance achievable through model combination.
Across several architectures and benchmarks, we find a substantial gap between individual models and their oracle, revealing that different models capture complementary knowledge. Yet, this complementarity rapidly saturates as more models are added, leaving a persistent subset of queries unsolved even by a large number of models. These findings reveal both the potential of model complementarity and a fundamental limit to what current link prediction models can collectively recover; thereby highlighting the need for further research to build robust Web applications.

\end{abstract}

\section{Introduction}

The Web exposes users to an ever-growing amount of information, making structuring, identifying, and ranking relevant information from large and heterogeneous collections of data a central challenge of modern Web systems (e.g., search, recommendation).
That is why, knowledge graphs (KGs) are increasingly used as a source of structured knowledge in Web applications, e.g. in semantic search~\cite{SemanticSearch}, question answering~\cite{question_answering,QA-GNN}, and recommender systems~\cite{recommender,RippleNet}.
KGs explicitly models entities and the relationships between them~\cite{hogan2021knowledge}, and are typically defined within standard frameworks such as RDF~\cite{Lanthaler:14:RCA}, where knowledge is represented as triples $(h,r,t)$, indicating that a
relation $r$ links a head entity $h$ to a tail entity $t$.

However, real-world KGs inherently incomplete~\cite{KrompassISWC2015}, limiting their usefulness and motivating several refinement tasks.
Among them, link prediction (LP) aims to infer missing entities in incomplete triples of the form $(h, r, ?)$ or $(?, r, t)$ within a learning-to-rank setting.
A wide range of methods have been developed for this task, including KG embedding models~\cite{TransE,DistMult,ConvE}, other neural approaches~\cite{minerva,NBFNet,NeuralLP}, and rule-based methods~\cite{AMIE,AnyBURL}.
Over the years, continuous improvements in model architectures and training strategies, such as loss design, regularization, and negative sampling, have led to improvements in performance as measured by metrics such as MRR~\cite{OldDog}.

Despite substantial progress in link prediction, individual models remain imperfect and, more importantly, do not necessarily capture the same knowledge.
Recent work~\cite{LinkPerdition,zhu2024multiplicty} has shown that training the same model with different random seeds can lead to different model \emph{instances} producing substantially different predictions.
This phenomenon has been quantified through metrics such as ambiguity~\cite{zhu2024multiplicty} and prediction overlap~\cite{LinkPerdition}, with results respectively showcasing that some instances may successfully rank the ground-truth entity among their top-K predictions while others fail to do so, and that a substantial portion of the top-K predictions may differ.

This suggests that different models and model instances may capture complementary parts of the knowledge that can be inferred from a KG, making model combination a potential means of increasing knowledge coverage. This intuition is consistent with the broader literature on ensemble learning for link prediction, where combining multiple predictors can exploit complementary predictions and improve performance~\cite{Krompass2015EnsembleSF,XuNVL21,GregucciN0S23}.
However, despite these improvements, ensemble methods remain far from perfect performance. 
Besides, to the best of our knowledge, no prior work has quantified how much additional performance could theoretically be obtained by fully exploiting the diversity of predictions across instances and models, nor how this potential evolves as an increasing number of instances is considered. This motivates the following research questions aiming at assessing model complementarity through performance gains:

\begin{description}
    \item[RQ1.] To what extent can multiple instances of a single model provide performance gains over a single instance?

    \item[RQ2.] To what extent can different models provide performance gains beyond those obtained from multiple instances of a single model?

    \item[RQ3.] How does the performance gain evolve as the number of instances increases?

\end{description}

To address these questions, we introduce an oracle framework, defined as an idealized setting that selects, for each test triple, the best prediction among a set of instances. This framework provides \emph{an upper bound} on the performance achievable through instance selection and allows us to quantify the potential performance gains of ensemble methods.

With this oracle framework, we identify three distinct performance gaps.
For RQ1, the \emph{Instance--Oracle Gap} highlights that exploiting seed-induced diversity alone is sufficient to yield substantial gains over individual instances.
For RQ2, the \emph{Model--Cross-model Gap} shows that combining different model architectures provides additional advantages, as cross-model ensembles reach the performance of the best single-model oracle without prior model selection.
Finally, for RQ3, the \emph{Asymptotic Gap} captures the performance levels that remain unattainable as more instances are added. Indeed, oracle performance quickly saturates, leaving a subset of queries that none of the considered models can solve.

The remainder of this paper is organized as follows. We first review related work in Section~\ref{sec:related}, before formalizing our oracle framework in Section~\ref{sec:oracle} and describing the experimental setup in Section~\ref{sec:expe}. Section~\ref{sec:result} then presents the results, structured around the three research questions and their corresponding performance gaps. Section~\ref{sec:discussion} discusses the findings, while Section~\ref{sec:conclusion} concludes the paper.

\section{Related Work}
\label{sec:related}
\subsection{Link Prediction Models}

Link prediction is a fundamental task for completing knowledge graphs
and supporting downstream applications such as search, question
answering, and recommendation~\cite{QA-GNN,RippleNet}.

Models for link prediction can be unified under the common formulation of a scoring function $\varphi(h, r, t)$, which evaluates the plausibility of a triple $(h, r, t)$. 

Over the past decade, the dominant approach has been KGEMs, which aim to embed a KG within a $d$-dimensional continuous vector space while preserving its structural regularities. 
The scoring function operates over this embedding space, which is learned during training by maximizing the scores of observed triples while minimizing those of negative ones generated by corrupting positive triples.
KGEMs are commonly categorized into three families based on the design of their scoring functions: geometric models (e.g., TransE~\cite{TransE}, RotatE~\cite{RotatE}, BoxE~\cite{BoxE}), tensor factorization models (e.g., RESCAL~\cite{RESCAL}, DistMult~\cite{DistMult}, ComplEx~\cite{ComplEx}), and neural network models that learn the scoring function (e.g., ConvE~\cite{ConvE}, HittER~\cite{Hitter}). 

Beyond the KGEMs introduced above, several approaches adopt distinct mechanisms while still relying on learned parameters.
NBFNet \cite{NBFNet} formulates the task as a path search problem, generalising the Bellman-Ford algorithm within a graph neural network. 
MINERVA~\cite{minerva} models the task as a sequential decision process, where an agent learns to navigate the knowledge graph by following relation paths from a source entity to a target entity, using reinforcement learning.
Alternatively, NeuralLP \cite{NeuralLP} compiles logical inference into differentiable tensor operations, enabling the joint learning of parameters and logical rule structures through gradient descent.

Alternatively to these neural approaches, symbolic rule mining methods extract explicit patterns directly from the graph.
Among them, AMIE~\cite{AMIE} follows a top-down search guided by confidence measures, while AnyBURL~\cite{AnyBURL} adopts a bottom-up strategy based on random walks. 

\subsection{Ensemble Learning for Link Prediction}
Recent work has explored ensemble learning strategies to improve link prediction performance. Xu et al.~\cite{XuNVL21} study ensembles composed of multiple instances of the same model and show that variability induced by different random seeds is sufficient to outperform a single model with an equivalent number of parameters. In particular, they exhibit that such ensembles can collectively capture relational patterns that are theoretically inaccessible to individual instances, such as symmetry in TransE. SnapE~\cite{SnapE} extends this idea by constructing ensembles from intermediate training snapshots, aiming to retain the benefits of ensembling without increasing training time.

These observations relate to the fact that KGEMs differ in the relational patterns they can theoretically capture (e.g., symmetry, intersection, composition), as a direct consequence of their scoring functions~\cite{RotatE,BoxE}. This has motivated ensemble approaches that aim to overcome the limitations of individual models by combining their strengths. Krompass and Tresp~\cite{Krompass2015EnsembleSF}, Gregucci et al.~\cite{GregucciN0S23}, and Yue et al.~\cite{YueZYLWZL023} argue that such combinations lead to improved predictive performance. More broadly, Meilicke et al.~\cite{RuleAndEmbedding,NaiveNeuroSymbo} show that KGEMs and rule-based methods can also effectively complement each other.

While these studies consistently report gains over the best single instance, the improvements in terms of MRR remain relatively modest~\cite{GregucciN0S23,YueZYLWZL023}. Moreover, this body of work leaves open the questions of a potential upper bound in performance when aggregating model instances, whether this upper bound is achievable, and whether existing approaches already operate close to this limit.
This motivates our proposed oracle framework to characterize such an upper bound, and measure the gaps between individual model instances and their oracle, opening further research on model aggregation strategies. 

\section{Measuring Performance Upper Bounds with Oracles}
\label{sec:oracle}

\begin{figure}
    \centering
    \includegraphics[width=1\linewidth]{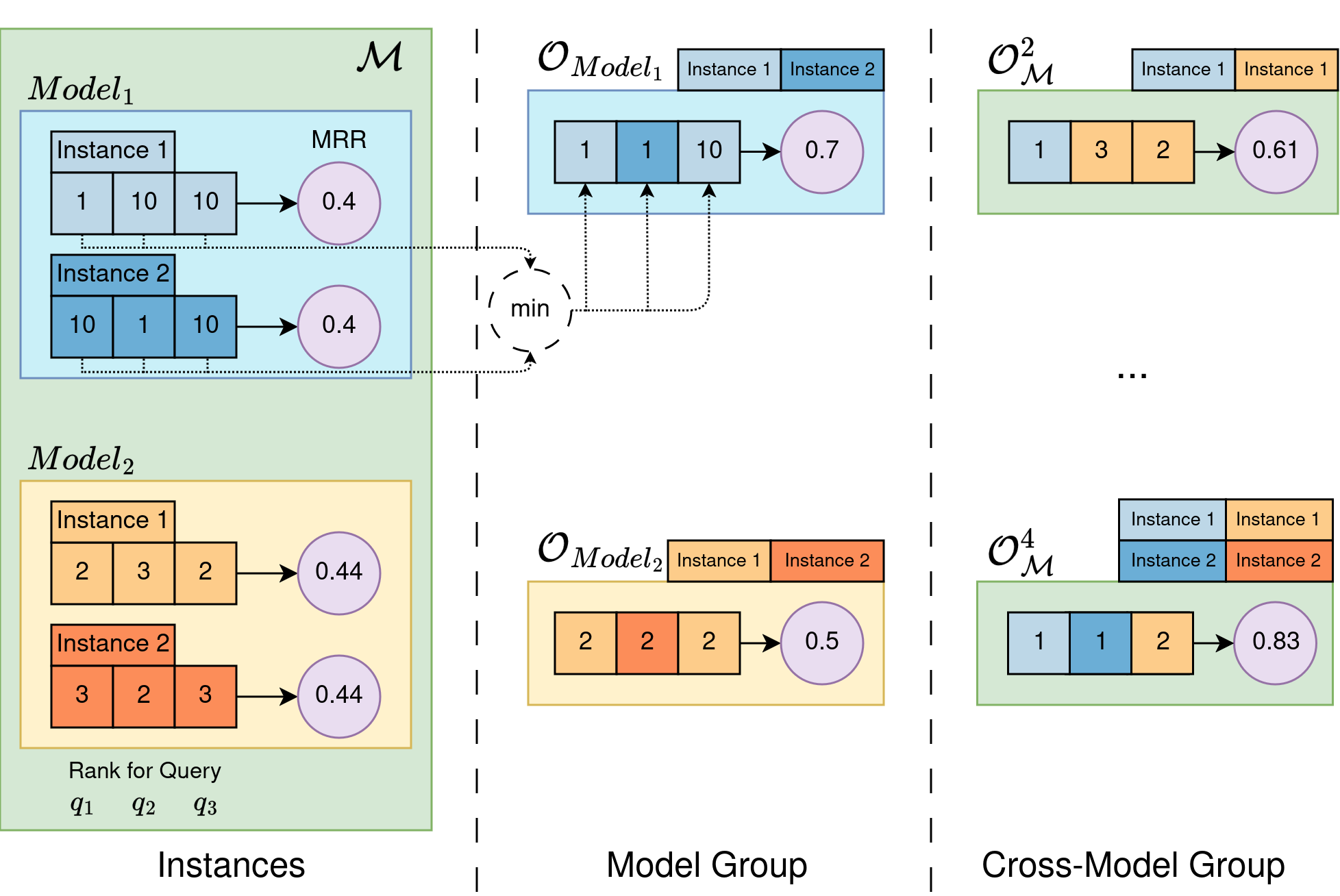}
    \caption{Per-query selection mechanism of oracles. Although individual instances achieve similar aggregate MRR, their performance differs at the query level ($q_1, q_2, q_3$). By selecting the best rank per query across instances, the oracle constructs optimal pseudo-instances at the model group level ($\mathcal{O}_{\text{Model}}$) and cross-model group level ($\mathcal{O}_{\mathcal{M}}$), leading to higher overall performance.}
    \label{fig:Fig1}
\end{figure}

Let us denote a KG as $\mathcal{K}=(\mathcal{E}, \mathcal{R}, \mathcal{T})$, where $\mathcal{E}$ is a set of entities, $\mathcal{R}$ a set of relations, and $\mathcal{T} \subseteq \mathcal{E} \times \mathcal{R} \times \mathcal{E}$ a set of triples encoding facts.
Each triple $(h, r, t)$ consists of a head entity $h$, a relation $r$, and a tail entity $t$, \textit{e.g.}, $(\text{France}, \text{hasCapital}, \text{Paris})$.
In link prediction, $\mathcal{T}$ is partitioned into fixed training ($\ttrain$), validation ($\tval$), and test ($\ttest$) sets. 
Given a test triple $(h,r,t)\in\ttest$, the task consists in predicting the missing entity in queries of the form $(h,r,?)$ or $(?,r,t)$ by scoring and ranking all entities, such that the ground-truth entity is ranked as highly as possible. 
Performance is thus evaluated using rank-based metrics such as Mean Rank (MR), Mean Reciprocal Rank (MRR), and Hits@$K$ (typically $K \in \{1,3,10\}$).

In the literature, several models have been proposed to address this task; in this work, we restrict our study to a subset of models denoted $\mathcal{M}$. 
The training of a model depends on the chosen architecture, a hyperparameter configuration that is often chosen as the one maximizing the rank-based metrics on the validation set, and a random seed to control stochastic factors such as negative sampling or embedding initialization. 
We call a trained model given these characteristics an \emph{instance}, that we formally define as follows:
\begin{definition}[Instance]

An instance $I = Instance(M, \mathcal{T}, \boldsymbol{C}, \mathfrak{S})$ corresponds to a model $M \in \mathcal{M}$ trained on dataset $\mathcal{T}$ with hyperparameter configuration $\boldsymbol{C}$ and random seed $\mathfrak{S}$.
When unambiguous, we write $I_M^{\mathfrak{S}}$.
\end{definition}

Even with $\mathcal{M}$, $\mathcal{T}$, and $\boldsymbol{C}$ fixed, training is stochastic: the learned representations depend on the random seed $\mathfrak{S}$~\cite{LinkPerdition}.
In standard practice, different instances obtained with distinct seeds are commonly treated as interchangeable, since aggregate link prediction metrics usually exhibit only limited variation across runs~\cite{LinkPerdition}.
Consequently, experimental results are often reported from a single run or averaged over a few runs, and the specific influence of the seed is largely disregarded. 

However, prior work has shown that such instances can yield divergent predictions~\cite{LinkPerdition,zhu2024multiplicty}.
This phenomenon can be illustrated in Figure~\ref{fig:Fig1}, where two instances of Model~1 achieve the same MRR (0.4), yet differ at the query level: the first instance ranks the first query correctly while the second does not, and the opposite holds for the second query. 
As previously introduced, this variability motivates the use of an ensemble of models to benefit from the different knowledge captured by the different instances~\cite{XuNVL21}.

In our work, we further investigate the performance gain brought by such groups of models.
To address the two settings of in-model and cross-model complementarity, we consider groups of models that can involve instances of the same model (Definition~\ref{def:model-group}), or instances of different models (Definition~\ref{def:cross-model-group}), respectively.

\begin{definition}[Model Group]
\label{def:model-group}
Let $M \in \mathcal{M}$ be a model. A \textbf{model group} associated with $M$ is a finite set of independently trained instances of $M$:
\[
G_M = \{ I_M^{\mathfrak{S}_1}, \dots, I_M^{\mathfrak{S}_l} \},
\]
where each instance $I_M^{\mathfrak{S}_i}$ is obtained using a different random seed $\mathfrak{S}_i$.
\end{definition}

\begin{definition}[Cross-Model Group]
\label{def:cross-model-group}
Let $\mathcal{M}' \subseteq \mathcal{M}$ be a set of models. A \textbf{cross-model group} is any finite set of instances of models in $\mathcal{M}'$:
\[
G_{\mathcal{M}'} = \{ I_{M_1}^{\mathfrak{S}_1}, \dots, I_{M_n}^{\mathfrak{S}_n} \},
\quad \text{with } M_i \in \mathcal{M}' \ \forall i,
\]
\end{definition}

To characterize the theoretical performance upper-bound of such groups, we model their theoretical best performing behavior as an \emph{oracle} formally defined as follows.
\begin{definition}[Oracle]
Let us denote $\mathcal{Q}$ the set of queries. Given a group $G$, the \textbf{oracle} associated with $G$ is the mapping
\[
  \begin{aligned}
    \mathcal{O}_{G} : \mathcal{Q} & \longrightarrow \mathbb{Z}^{+} \\
    q & \longmapsto \mathcal{O}_{G}(q)=\min_{I \in G} rank(I, q).
  \end{aligned}
\]
where $rank(I, q)$ returns the rank assigned by instance $I$ to the ground-truth entity for a query $q$.
\end{definition}
\noindent In other words, the oracle can be interpreted as a pseudo-instance that, for each query, achieves the best rank obtained by the instances in $G$.
For clarity, we denote the oracle of a model group $G_M$ or a cross-model group $G_{\mathcal{M}'}$ as $\mathcal{O}_M \equiv \mathcal{O}_{G_M}$ and $\mathcal{O}_{\mathcal{M}'} \equiv \mathcal{O}_{G_{\mathcal{M}'}}$, respectively.

Figure~\ref{fig:Fig1} provides examples of oracle predictions for Model~1. For the first query, the oracle selects the first instance, which ranks the ground-truth entity higher than the second instance ($1$ vs.\ $10$). The opposite occurs for the second query. For the third query, both instances yield identical ranks, and the oracle selection is therefore arbitrary. The same procedure applies to Model~2.
Figure~\ref{fig:Fig1} also illustrates cross-model group oracles with 2 and 4 instances per group. The selection mechanism remains identical, but the oracle can now choose among instances from different models. For example, $\mathcal{O}^4_{\mathcal{M}}$ will first select instance~1 and then instance~2 from Model~1, and then select the prediction of instance~1 from Model~2, which ranks higher than the Model~1 instances ($2$ vs.\ $10$).

This pseudo-instance interpretation allows us to evaluate $\mathcal{O}_G$ using the same metrics as for standard instances.
In particular, the Mean Reciprocal Rank of the oracle is defined as
\[
\mathrm{MRR}_{\mathcal{O}_{G}}
=
\frac{1}{|\mathcal{Q}|}
\sum_{q \in \mathcal{Q}}
\frac{1}{\mathcal{O}_{G}(q)}.
\]
\noindent $\mathrm{MRR}_{\mathcal{O}_{G}}$ therefore represents the upper-bound performance achievable if, for each query, one could \emph{a priori} select the best-performing instance in $G$.
This upper-bound thus characterizes the theoretical performance limit of a group of instances w.r.t. perfect prediction, and enables to assess the gap between instances and this upper-bound, paving the way toward enhanced strategies for result aggregation.

To illustrate, in Figure~\ref{fig:Fig1}, although the instances of Model~1 and Model~2 individually obtain MRR values of $0.40$ and $0.44$, respectively, the collective performance measured by the oracle is higher. In particular, two instances of Model~1 can jointly achieve an MRR of $0.70$, while Model~2 reaches $0.50$.
When combining the first two instances of Model~1 and Model~2, the resulting group achieves an MRR of $0.61$, and this value further increases to $0.83$ when considering the full set of instances.

\section{Experimental Settings}

\label{sec:expe}

\paragraph{Models.}
\label{sec:models}

In our experiments, we consider a diverse set of representative models for link prediction, spanning several methodological families.
First, we include a collection of widely adopted knowledge graph embedding models:
\begin{multline*}
\mathcal{M}_{\text{KGE}} = \{\text{TransE}, \text{ComplEx}, \text{DistMult}, \text{ConvE}, \text{RotatE}, \text{RESCAL}, \\ \text{BoxE}\}.
\end{multline*}

We then extend this set with neural approaches relying on alternative formalisms, such as path-based reasoning using reinforcement learning, GNN, or neural rule induction:
\begin{multline*}
\mathcal{M}_{\text{Neural}} = \mathcal{M}_{\text{KGE}} \cup \{\text{NeuralLP}, \text{NBFNet}, \text{MINERVA}\}.
\end{multline*}

In addition, we consider two symbolic rule-based methods for link prediction:
\[
\mathcal{M}_{\text{Symbolic}} = \{\text{AMIE}, \text{AnyBURL}\}.
\]

Finally, the complete set of models considered in this work is defined as:
\[
\mathcal{M} = \mathcal{M}_{\text{Neural}} \cup \mathcal{M}_{\text{Symbolic}}.
\]

For model implementation, we rely on reference frameworks and implementations, depending on model availability and in decreasing order of priority: LibKGE~\cite{libkge} (for TransE, DistMult, ComplEx, ConvE, RotatE, and RESCAL), PyKEEN~\cite{pykeen} (for BoxE), 
TorchDrug\footnote{\url{https://github.com/DeepGraphLearning/torchdrug}} (for NBFNet and NeuralLP), PyClause~\cite{pyclause} (for AMIE and AnyBURL). 
Finally, we use the original implementation of MINERVA~\cite{minerva}, as it is not available in the aforementioned frameworks.
Note that the original implementation of \text{MINERVA} trains exclusively for tail prediction. To allow comparison with the other considered models, we choose to report results for both head and tail prediction. Specifically, we reformulate each head-prediction query $(?,r,t)$ as the equivalent tail-prediction query $(t,r^{-1},?)$.
This formulation for head prediction is valid since \text{MINERVA} learns to navigate the graph using both relations and their inverses.
Also note that AMIE is included in cross-model comparisons, but differs fundamentally from the other approaches as it relies on a fully deterministic rule mining process.

\paragraph{Datasets.}
\label{sec:dataset}

We consider three standard datasets from the literature: \textbf{WN18RR}~\cite{ConvE}, a subset of the lexical KG WordNet characterized by its hierarchical structure; and \textbf{FB15k-237}~\cite{FB15k-237} and \textbf{CoDEx-S}~\cite{codex}, derived from Freebase and Wikidata, respectively, two prominent Web KGs containing large collections of real-world facts.
We use the standard splits of these datasets, with PyKEEN's filtering to remove test triples involving entities unseen during training, which removes $28$ out of $20,466$ test triples from FB15k-237, $210$ out of $3,134$ from WN18RR, and none from CoDEx-S.

\paragraph{Instances.}

For each $(\text{model}, \text{dataset})$ pair, we train $20$ independent instances using random seeds ranging from $0$ to $19$.
Instances that fail to converge, defined as achieving a validation MRR strictly below $0.05$, are discarded from the analysis. Consequently, we exclude $1$, $9$, and $9$ NeuralLP instances on WN18RR, FB15k-237, and CoDEx-S, respectively.

For TransE, DistMult, ComplEx, ConvE, RotatE, and RESCAL, we use the official configurations provided in the LibKGE repository for FB15k-237 and WN18RR. These configurations result from the hyperparameter tuning procedure conducted by Ruffinelli et al.~\cite{OldDog}. For CoDEx-S, we use the configurations provided in the same repository for TransE, ComplEx, and ConvE, obtained following the hyperparameter tuning procedure as in Safavi et al.~\cite{codex}. As no dataset-specific configurations are provided for DistMult and RotatE on CoDEx-S, we reuse their FB15k-237 configurations, given the similarity between the two datasets.

For BoxE, MINERVA, and NBFNet, we follow the same principle. We use the official configurations provided by their respective authors for FB15k-237 and WN18RR~\cite{BoxE,minerva,NBFNet}, and reuse the FB15k-237 configurations for CoDEx-S. For NeuralLP, we use the same configuration across all datasets, following the configuration provided in the official TorchDrug tutorial\footnote{\url{https://torchdrug.ai/docs/tutorials/reasoning.html}}.

Finally, AnyBURL and AMIE, implemented with PyClause, are run using the framework's default configurations.

\paragraph{Evaluation Metrics.} To evaluate instance performance, we focus on the standard Mean Reciprocal Rank (MRR), Hit@1 and Hit@10 metrics. We use \emph{filtered} ranks, removing all other known true triples from the candidate set, and the \emph{pessimistic} tie convention, assigning tied candidates the worst rank in their tie group.

\paragraph{Experimental Protocols.}
\label{sec:exp}
We devise specific experimental protocols to answer each of the research questions.

\textbf{RQ1 (Section~\ref{sec:gap}).}
For each model $M \in \mathcal{M}$, we consider its associated group $G_M$ composed of $20$ independently trained instances. We compare the average performance across instances, $\overline{\mathrm{MRR}}_{G_M}$, with the oracle performance over the group, $\mathrm{MRR}_{\mathcal{O}_M}$.

\textbf{RQ2 (Section~\ref{sec:complem}).}
We evaluate the performance of cross-model groups. To ensure a fair comparison with single-model groups, we control the total number of instances. Specifically, we sample the same number of instances from each model in a given family $\mathcal{M}'$, such that the total number of selected instances $k$ remains close to $20$. The corresponding oracle performance is denoted $\mathrm{MRR}^{k}_{\mathcal{O}_{\mathcal{M}'}}$. As this sampling is stochastic, we repeat the process $100$ times and report the average oracle performance.

To relax the uniform allocation constraint, we construct a group of $20$ instances using a greedy procedure over $\mathcal{M}$, iteratively selecting the instance that yields the largest increase in oracle performance, regardless of its model. Once the greedy procedure is completed, we repeatedly sample new instances while preserving the model proportions of the selected group. This allows us to disentangle the effect of the resulting model allocation from that of the particular instances selected during the greedy procedure, and thus assess whether the observed gain is attributable to a promising model allocation rather than to specific instances. We perform $100$ such samplings and report the resulting average oracle performance as $\mathrm{MRR}^{20,\text{greedy}}_{\mathcal{O}_{\mathcal{M}}}$.

\textbf{RQ3 (Section~\ref{sec:dynamic}).}
We analyze how oracle performance evolves as the number of instances increases. Given the instances of a group $G$, we evaluate $\mathrm{MRR}^{k}_{\mathcal{O}_G}$ for $k \in \{1, \dots, |G|\}$ by progressively considering subsets of size $k$. Since the results depend on the order in which instances are added, we sample $100$ random permutations and report the average oracle performance for each value of $k$.

Our code is available on GitHub.\footnote{\label{footnote:github}\url{https://anonymous.4open.science/r/OracleWill_I_Learn/README.md}}

\section{Uncovering Performance Gaps}
\label{sec:result}

Our objective is to quantify how much can be gained by moving from a single instance to groups of instances and models. To this end, we structure our analysis around three gaps. We first study the \emph{Instance--Oracle Gap}, measuring how much a group of instances of the same model can collectively outperform a single instance (Section~\ref{sec:gap}). We then examine the \emph{Model--Cross-Model Gap}, capturing the additional gains brought by combining different models (Section~\ref{sec:complem}). Finally, we analyze the \emph{Asymptotic Gap}, highlighting that performance converges below perfect accuracy as the number of instances increases (Section~\ref{sec:dynamic}).

\subsection{Instance--Oracle Gap: Performances of Instances Versus Their Oracle}
\label{sec:gap}

Given a model $M$, we compare the average MRR of the instances within a group, $\overline{\mathrm{MRR}}_{G_M}$, with the MRR of the oracle of the same group, $\mathrm{MRR}_{\mathcal{O}_{M}}$.   
This enables to assess the impact of seed-induced variability on performance.
Results are reported in Table~\ref{tab:rq1_mrr} (and in Appendix~\ref{app:reshit} for Hit@1).
When aggregating individual MRR scores, the standard deviation across seeds is close to zero for most models. 
Such performance stability could suggest limited diversity across runs and therefore negligible room for improvement beyond single-instance performance.
However, substantial gaps consistently appear between $\overline{\mathrm{MRR}}_{G_{M}}$ and $\mathrm{MRR}_{\mathcal{O}_{M}}$, revealing a markedly different picture. 
For example, on CoDEx-S, ConvE improves from $0.45$ to $0.64$, DistMult from $0.42$ to $0.56$, and ComplEx from $0.46$ to $0.60$. Across all datasets, these gains often exceed $10$ points and can reach up to nearly $20$ points, demonstrating that instances trained with different seeds solve partially disjoint subsets of queries.

Moreover, although strong instance-level performance tends to be associated with higher oracle performance, this relationship does not always hold. For instance, on CoDEx-S, ComplEx achieves a higher averaged MRR than ConvE ($0.46$ vs.\ $0.45$), yet exhibits a lower oracle performance ($0.60$ vs.\ $0.64$). This suggests that oracle gains depend not only on instance-level performance alone but also on the diversity of correctly predicted queries across instances.

\begin{table}
\centering
\caption[]{Averaged and Oracle MRR across datasets, model groups and cross-model groups.}
\small
\setlength{\tabcolsep}{3pt}
\begin{tabular}{cc lccc}
\toprule
 & & Model & WN18RR & FB15k-237 & CoDEx-S \\
\midrule

\multirow{20}{*}{\rotatebox{90}{\textbf{Neural}}} 
& \multirow{14}{*}{\rotatebox{90}{KGE}}
& $\overline{\mathrm{MRR}}_{G_{\text{TransE}}}$ & $.24 \pm .00$ & $.31 \pm .00$ & $.35 \pm .00$ \\
& & $\mathrm{MRR}_{\mathcal{O}_{\text{TransE}}}$ & $.27$ & $.45$ & $.43$ \\
& & $\overline{\mathrm{MRR}}_{G_{\text{DistMult}}}$ & $.48 \pm .00$ & $.34 \pm .00$ & $.42 \pm .00$ \\
& & $\mathrm{MRR}_{\mathcal{O}_{\text{DistMult}}}$ & $.53$ & $.46$ & $.56$ \\
& & $\overline{\mathrm{MRR}}_{G_{\text{ConvE}}}$ & $.47 \pm .00$ & $.34 \pm .00$ & $.45 \pm .00$ \\
& & $\mathrm{MRR}_{\mathcal{O}_{\text{ConvE}}}$ & $.54$ & $.47$ & $.64$ \\
& & $\overline{\mathrm{MRR}}_{G_{\text{RotatE}}}$ & $.51 \pm .00$ & $.33 \pm .00$ & $.42 \pm .00$ \\
& & $\mathrm{MRR}_{\mathcal{O}_{\text{RotatE}}}$ & $.56$ & $.42$ & $.53$ \\
& & $\overline{\mathrm{MRR}}_{G_{\text{ComplEx}}}$ & $.51 \pm .00$ & $.35 \pm .00$ & $.46 \pm .00$ \\
& & $\mathrm{MRR}_{\mathcal{O}_{\text{ComplEx}}}$ & $.60$ & $.46$ & $.60$ \\
& & $\overline{\mathrm{MRR}}_{G_{\text{RESCAL}}}$ & $.50 \pm .00$ & $.36 \pm .00$ & $.40 \pm .00$ \\
& & $\mathrm{MRR}_{\mathcal{O}_{\text{RESCAL}}}$ & $.59$ & $.52$ & $.54$ \\
& & $\overline{\mathrm{MRR}}_{G_{\text{BoxE}}}$ & $.48 \pm .00$ & $.35 \pm .00$ & $.42 \pm .01$ \\
& & $\mathrm{MRR}_{\mathcal{O}_{\text{BoxE}}}$ & $.55$ & $.42$ & $.50$ \\
\cmidrule{2-6}

& & $\overline{\mathrm{MRR}}_{G_{\text{MINERVA}}}$ & $.43 \pm .01$ & $.17 \pm .00$ & $.29 \pm .01$ \\
& & $\mathrm{MRR}_{\mathcal{O}_{\text{MINERVA}}}$ & $.49$ & $.32$ & $.47$ \\
& & $\overline{\mathrm{MRR}}_{G_{\text{NeuralLP}}}$ & $.34 \pm .04$ & $.21 \pm .01$ & $.27 \pm .01$ \\
& & $\mathrm{MRR}_{\mathcal{O}_{\text{NeuralLP}}}$ & $.53$ & $.38$ & $.44$ \\
& & $\overline{\mathrm{MRR}}_{G_{\text{NBFNet}}}$ & $.59 \pm .00$ & $.41 \pm .00$ & $.36 \pm .01$ \\
& & $\mathrm{MRR}_{\mathcal{O}_{\text{NBFNet}}}$ & $.68$ & $.58$ & $.53$ \\
\midrule

\multirow{4}{*}{\rotatebox{90}{\textbf{Symbolic}}}
& & $\overline{\mathrm{MRR}}_{G_{\text{AnyBURL}}}$ & $.48 \pm .00$ & $.25 \pm .00$ & $.30 \pm .00$ \\
& & $\mathrm{MRR}_{\mathcal{O}_{\text{AnyBURL}}}$ & $.49$ & $.34$ & $.31$ \\
& & $\overline{\mathrm{MRR}}_{G_{\text{AMIE}}}$ & $.43 \pm .00$ & $.20 \pm .00$ & $.23 \pm .00$ \\
& & $\mathrm{MRR}_{\mathcal{O}_{\text{AMIE}}}$ & $.43$ & $.20$ & $.23$ \\
\midrule

\multirow{9}{*}{\rotatebox{90}{\textbf{Cross-model}}}
& & $\mathrm{MRR}^{21}_{\mathcal{O}_{KGE}}$ & $.62 \pm .00$ & $.51 \pm .00$ & $.63 \pm .00$ \\ \addlinespace
& & $\mathrm{MRR}^{20}_{\mathcal{O}_{Neural}}$ & $.67 \pm .00$ & $.58 \pm .00$ & $.66 \pm .00$ \\ \addlinespace
& & $\mathrm{MRR}^{20}_{\mathcal{O}_{Symbolic}}$ & $.49 \pm .00$ & $.33 \pm .00$ & $.32 \pm .00$ \\ \addlinespace
& & $\mathrm{MRR}^{24}_{\mathcal{O}_{\mathcal{M}}}$ & $.67 \pm .00$ & $.58 \pm .00$ & $.66 \pm .00$ \\ \addlinespace
& & $\mathrm{MRR}^{20, \text{greedy}}_{\mathcal{O}_{\mathcal{M}}}$ & $.69 \pm .00$ & $.60 \pm .00$ & $.67 \pm .00$ \\ \addlinespace
& & $\mathrm{MRR}^{240}_{\mathcal{O}_{\mathcal{M}}}$ & $.73$ & $.69$ & $.76$ \\

\bottomrule
\end{tabular}
\label{tab:rq1_mrr}
\end{table}

Our results thus reveal a clear performance gap between the predictive ability of a single instance and that of a collection of instances. 
The potential gain from exploiting this collection is substantially larger than what is achieved by current ensemble learning methods~\cite{GregucciN0S23,YueZYLWZL023}. 
This gap highlights the potential for improved aggregation and ensemble strategies that can better exploit the complementary predictions of different instances, motivating further work toward practical methods that approach the oracle's performance.

\subsection{Model--Cross-Model Gap: Benefits of Mixing Models}
\label{sec:complem}

Prior work on ensemble learning~\cite{GregucciN0S23} assumes that structural differences across models constitute a complementary source of diversity. 
Building on this observation, this section investigates the gains associated with cross-model grouping, as well as the complementarity between models.
Results are reported in the last section of Table~\ref{tab:rq1_mrr}. 

Cross-model groups generally achieve higher oracle performance than most model groups. 
However, they do not systematically outperform the best single-model oracle. For example, on FB15k-237, the oracle performance obtained with 20 instances of NBFNet is identical ($0.58$) to that obtained with 24 instances (three per model) drawn from $\mathcal{M}$. 
In some cases, cross-model oracles may underperform the best individual model group; for instance, ConvE achieves a higher oracle score than the $\mathcal{M}_{\text{KGE}}$ cross-model group on CoDEx-S ($0.64$ vs.\ $0.63$).

Nevertheless, cross-model group oracles tend to closely approach the best single-model oracle. 
Interestingly, the latter can vary across datasets: within $\mathcal{M}_{\text{KGE}}$, the best single-model oracle is ComplEx on WN18RR, RESCAL on FB15k-237, and ConvE on CoDEx-S. 
This observation highlights the practical relevance of cross-model groups.
Indeed, in practice, it is not possible to determine in advance which specific model will perform best on a given dataset. 
Adopting a cross-model approach thus allows to expect the best performance without searching for the best performing single model, under the assumption that oracle performance can be attained

We also observe a remarkably low variance in cross-model oracle scores despite the use of sampling. This suggests that, as with the stability of MRR across instances, diversity is uniformly distributed: instances appear largely interchangeable, and substituting one instance for another has limited impact on overall complementarity.

\begin{figure}[t]
    \centering
    
    \begin{subfigure}{0.33\linewidth}
        \centering
        \includegraphics[width=\linewidth]{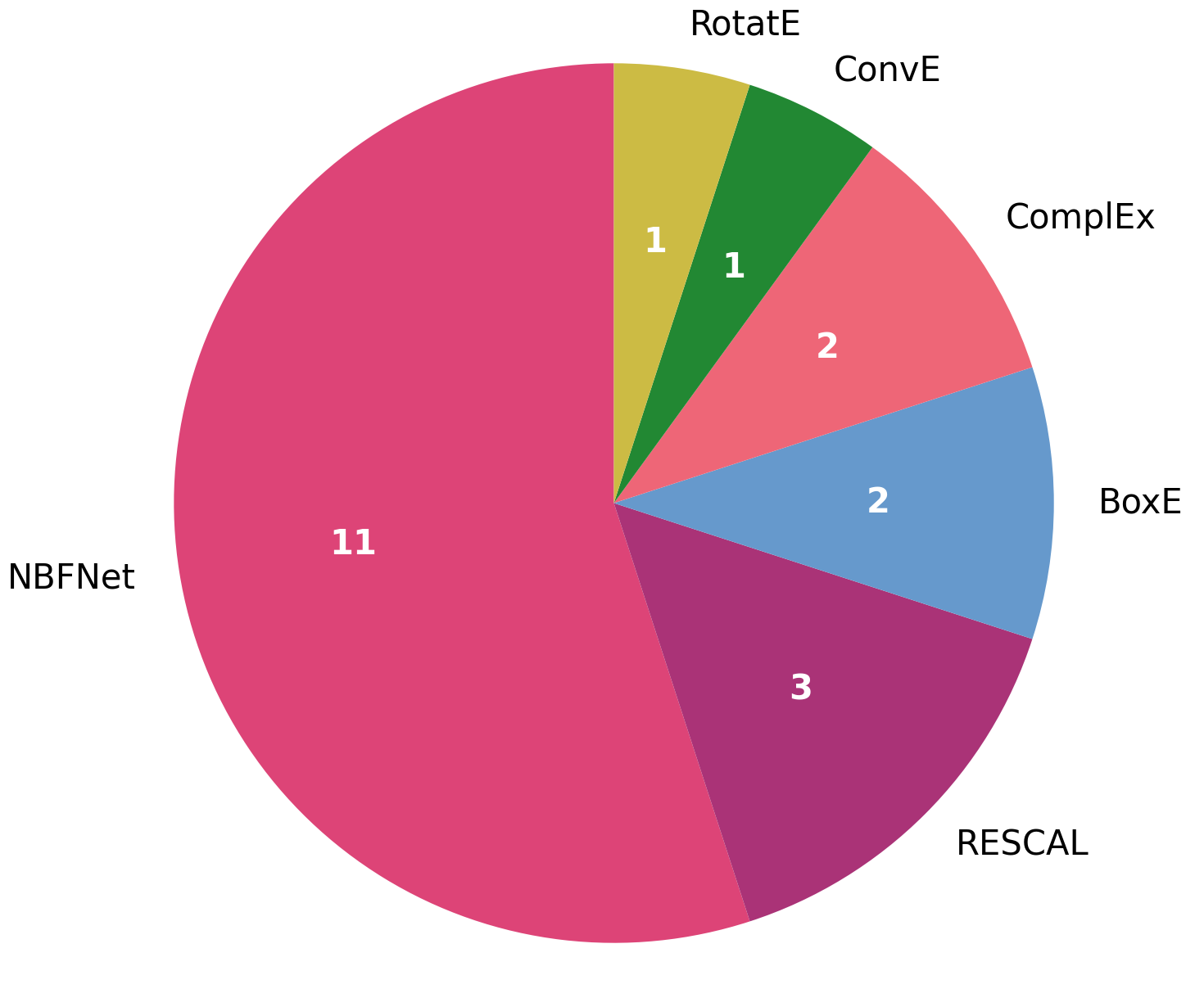}
        \caption{WN18RR}
    \end{subfigure}
    \hfill
    \begin{subfigure}{0.32\linewidth}
        \centering
        \includegraphics[width=\linewidth]{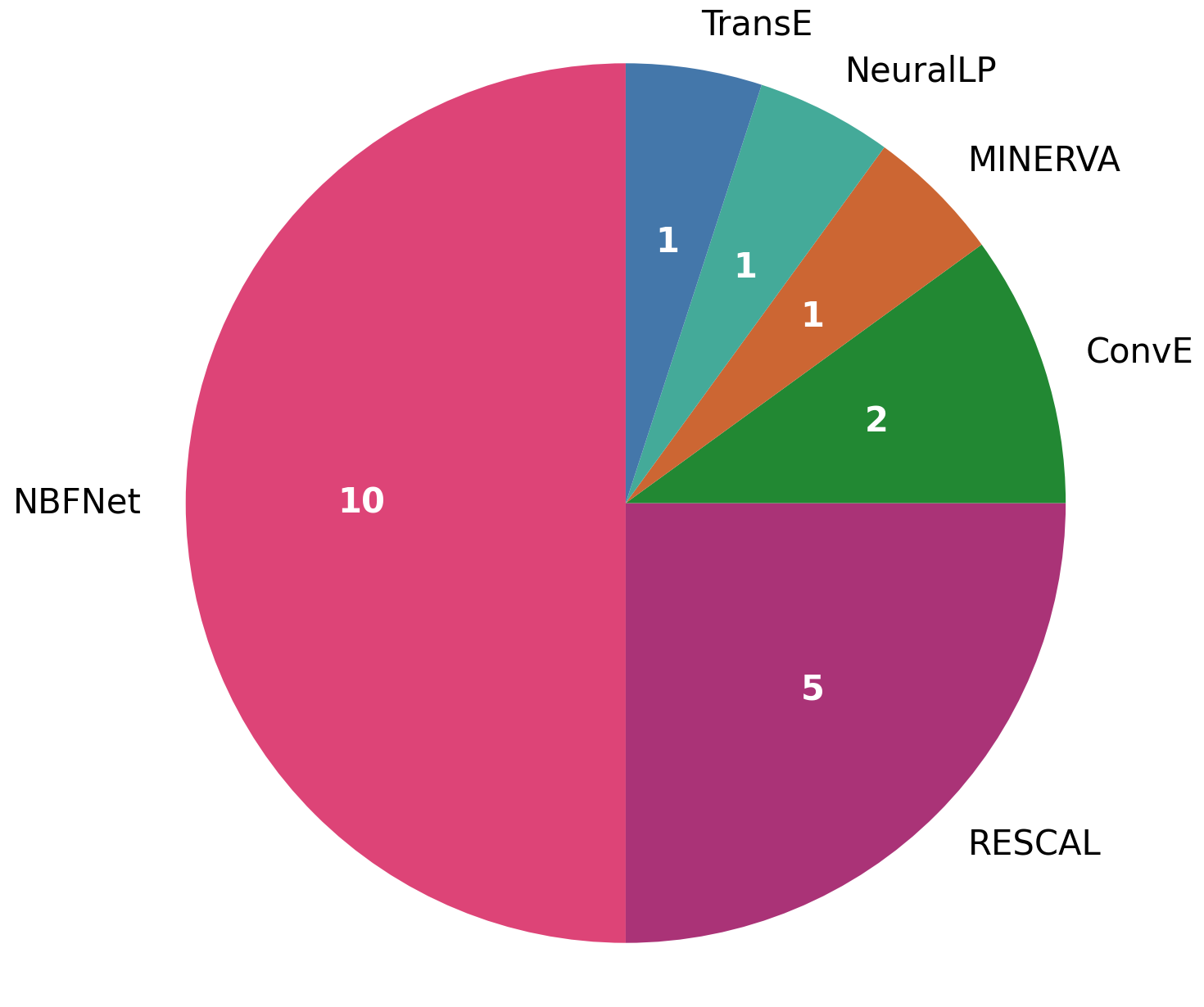}
        \caption{FB15k-237}
    \end{subfigure}
    \hfill
    \begin{subfigure}{0.32\linewidth}
        \centering
        \includegraphics[width=\linewidth]{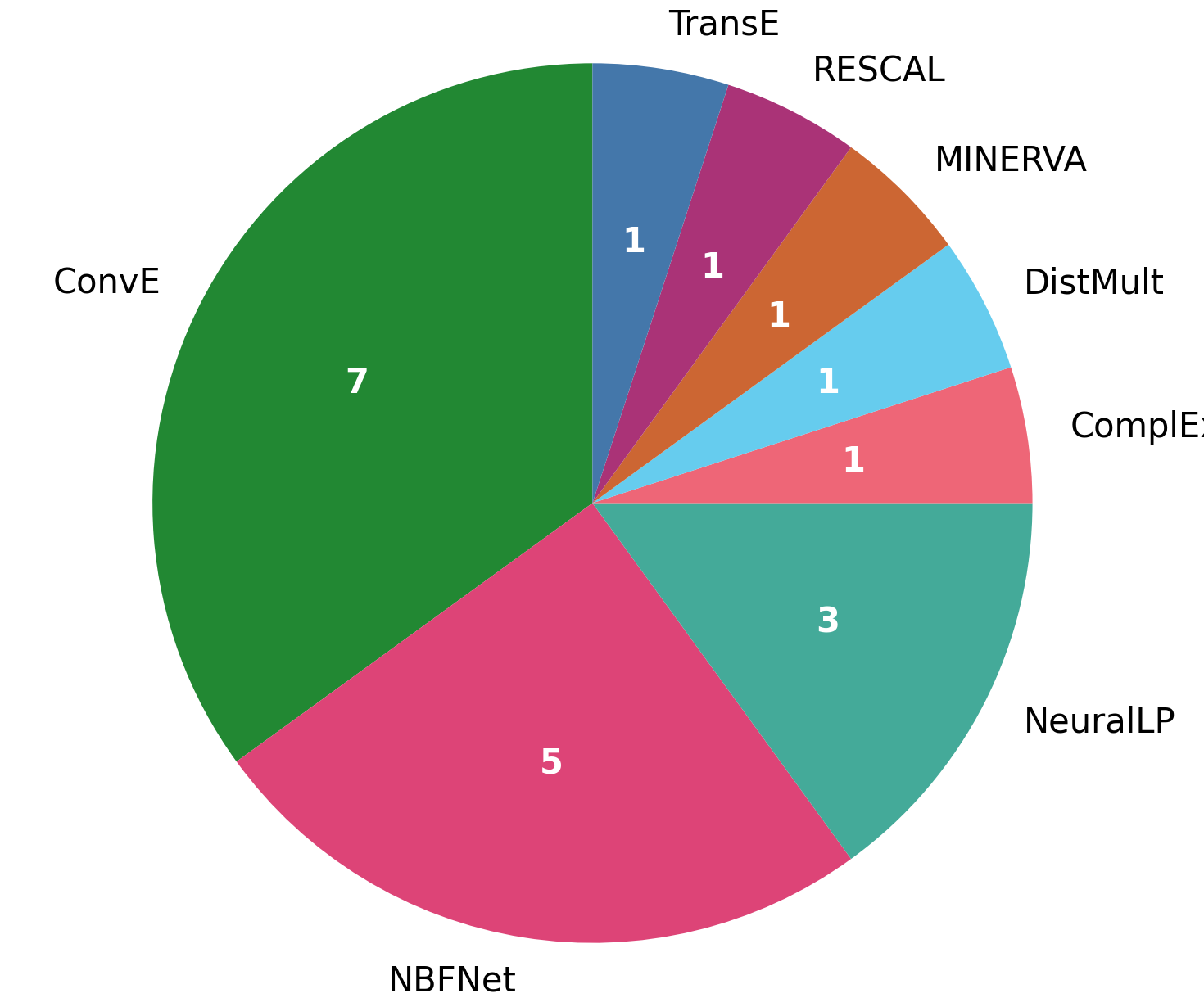}
        \caption{CoDEx-S}
    \end{subfigure}

    \caption{Greedy composition across datasets.}
    \label{fig:compo_greedy}

\end{figure}

Enforcing a uniform composition of models within cross-model groups may be restrictive. To relax this constraint, we consider $\mathrm{MRR}^{20, \text{greedy}}_{\mathcal{O}_{\mathcal{M}}}$, which constructs a group by greedily selecting instances so as to maximize oracle performance (see the experimental protocol for RQ2 in Section~\ref{sec:expe}). 
Across all datasets, the oracle of the resulting group consistently outperforms (albeit with a modest margin) the best single-model oracle ($0.69$ vs.\ $0.68$ on WN18RR, $0.60$ vs.\ $0.58$ on FB15k-237, and $0.67$ vs.\ $0.64$ on CoDEx-S).
This result confirms that cross-model groups benefit from model complementarity, while suggesting that a heterogeneous composition of models may bring additional performance gains, with models providing more diversity than others.
This phenomenon is illustrated in Figure~\ref{fig:compo_greedy}, which depicts the composition of the greedily constructed group.
A clear predominance of NBFNet emerges, consistent with its strong individual performance on FB15k-237 and WN18RR. 
However, other models, including lower-performing ones such as NeuralLP, TransE, and MINERVA, are also selected on FB15k-237. This highlights that performance is not sufficient to characterize the complementary contribution of a model to a group: a model with lower individual performance may still provide substantial complementarity by correctly predicting queries that are missed by stronger models.

\begin{figure}
    \centering
    \includegraphics[width=1\linewidth]{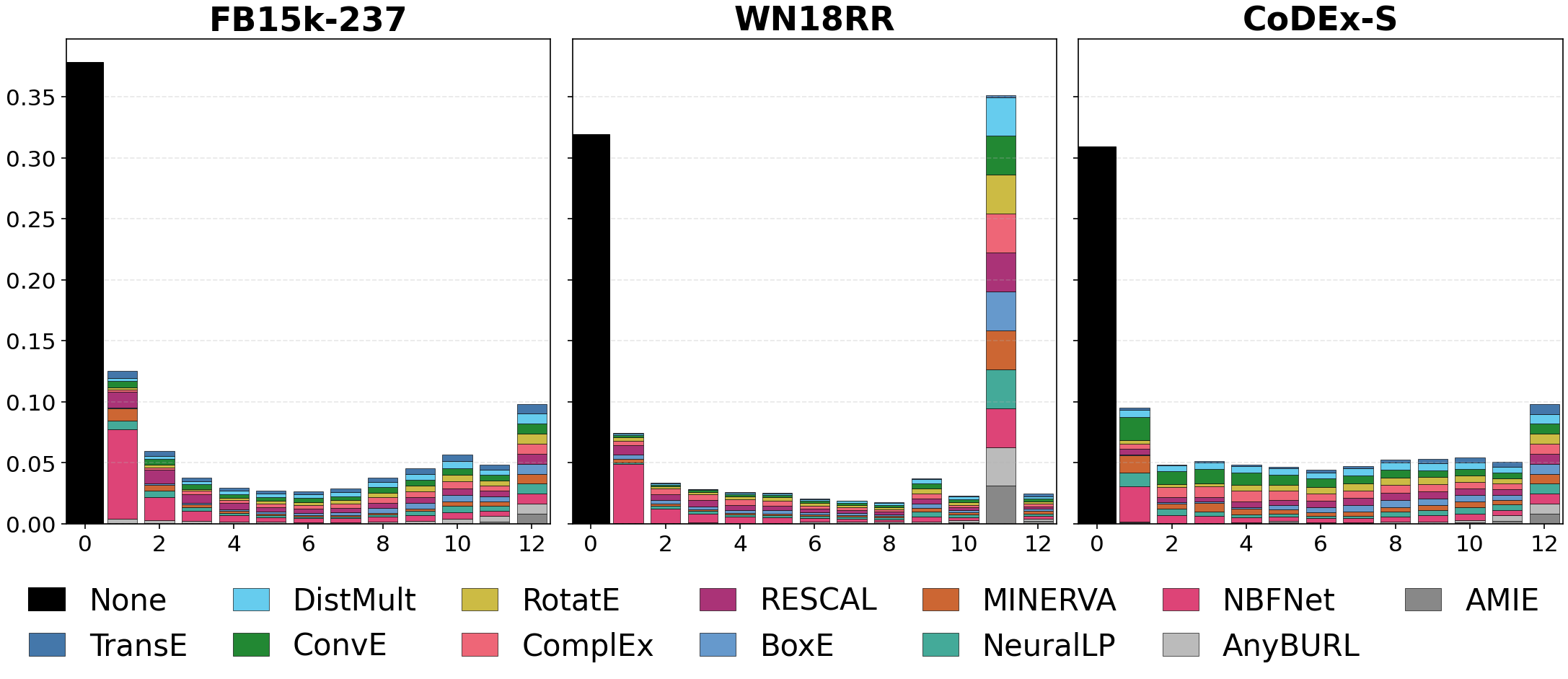}
    \caption{Distribution of queries according to the number of model oracles achieving Hit@1}
    \label{fig:count_hit_model}
\end{figure}

To better qualify this diversity, we analyze whether different model oracles succeed on the same queries. To this end, we reuse model-group oracles and binarize their predictions using Hit@1, indicating whether a given oracle correctly answers a query. For each query, we then count the number of model oracles that successfully resolve it, while also tracking the contribution of each model. The results are presented in Figure~\ref{fig:count_hit_model}.

\sloppy
First, we observe that a substantial fraction of queries, approximately one third across all datasets, are not solved by any model. Furthermore, the proportion of queries solved by only a small number of model oracles (one to three) is significant, accounting for nearly $20\%$ on FB15k-237 and CoDEx-S, and about $15\%$ on WN18RR. We also observe that WN18RR exhibits a bimodal distribution, with queries either unsolved by all models or solved by nearly all (except TransE).
Finally, the distribution further reveals that, although NBFNet dominates on FB15k-237 and WN18RR, other models also contribute in non-negligible proportions. 
A more fine-grained view of these interactions is provided by the upset plots in Appendix~\ref{app:upset}.
Overall, these observations confirm that, from an oracle perspective, complementarity can ensure strong performance, and even yield additional gains when sufficiently diverse groups are considered.

\begin{figure}[h]
    \centering
    \includegraphics[width=0.95\linewidth]{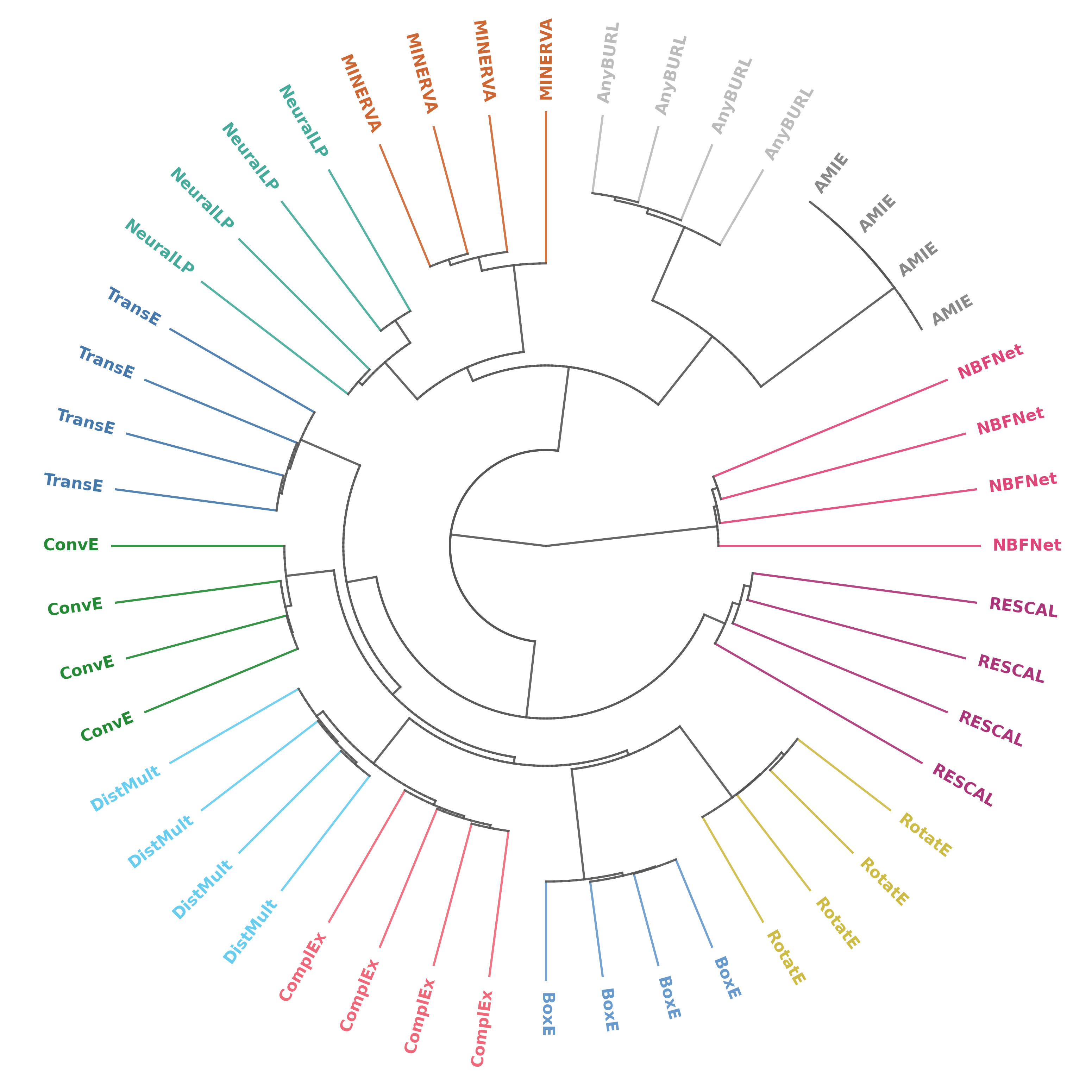}
    \caption{Dendrogram of instance similarities on FB15k-237.}
    \label{fig:FBdendojaccard}
\end{figure}

The previous results motivated an exploratory analysis of diversity across instances.
More precisely, we measure the redundancy between instances as the Jaccard distance computed over the sets of queries for which each instance achieves a Hit@1 of 1 on FB15k-237.
We then visualize this redundancy using a dendrogram (Figure~\ref{fig:FBdendojaccard}), where the distance between two instances reflects the similarity of their predictions.
Instances merging closer to leafs are thus more similar. 
Pairwise distances between instances are computed once, followed by agglomerative hierarchical clustering (UPGMA, average linkage). 

This visualization highlights several key patterns. First, instances from the same model form tight clusters, indicating that identical architectures tend to learn similar queries. Second, NBFNet clearly separates from other models, suggesting that it captures distinct queries. The remaining branch further splits into two groups: one containing the full $\mathcal{M}_{\text{KGE}}$ family, where DistMult and its generalization ComplEx appear particularly close, and a second cluster mixing symbolic and neural models. 
The latter further divides into two subgroups: one corresponding to $\mathcal{M}_{\text{Symbolic}}$, where AMIE exhibits no variation across instances, and another grouping NeuralLP and MINERVA. The proximity of NeuralLP to symbolic models can be explained by its joint learning of embeddings and rules, while the position of MINERVA suggests that path-based reasoning shares similarities with rule-based approaches.

The observation that the distance between two NBFNet instances is comparable to that between instances of different KGE models, or even between instances from the cluster mixing symbolic and neural approaches, indicates substantial diversity among NBFNet instances.
This diversity may explain why NBFNet accounts for a large proportion of the models selected by the greedy procedure, in addition to its strong individual performance. A similar trend is observed for RESCAL within the KGE family, where it appears to be among the most diversified models.

In this section, we have shown that cross-model grouping provides a reliable way to approach the best single-model oracle, without requiring dataset-specific model selection. 
We further observed that models exhibit some degree of complementarity in their predictions, which can be leveraged to yield additional performance gains. This complementarity therefore provides a natural basis for developing ensemble methods that exploit the diversity of predictions across different model architectures. 
A further research avenue lies in identifying \emph{a priori} highly complementary groups.

\subsection{Asymptotic Gap: Performance Limit under Large Groups}
\label{sec:dynamic}

Until now, our analysis has focused on groups with a fixed and limited number of instances, ensuring comparability across models and groups. However, the behavior of $\mathrm{MRR}^{240}_{\mathcal{O}_{\mathcal{M}}}$ suggests that increasing the number of instances can further improve performance, raising the question of how oracle performance evolves as the group size grows.

\begin{figure}[h]
    \centering
    \includegraphics[width=\linewidth]{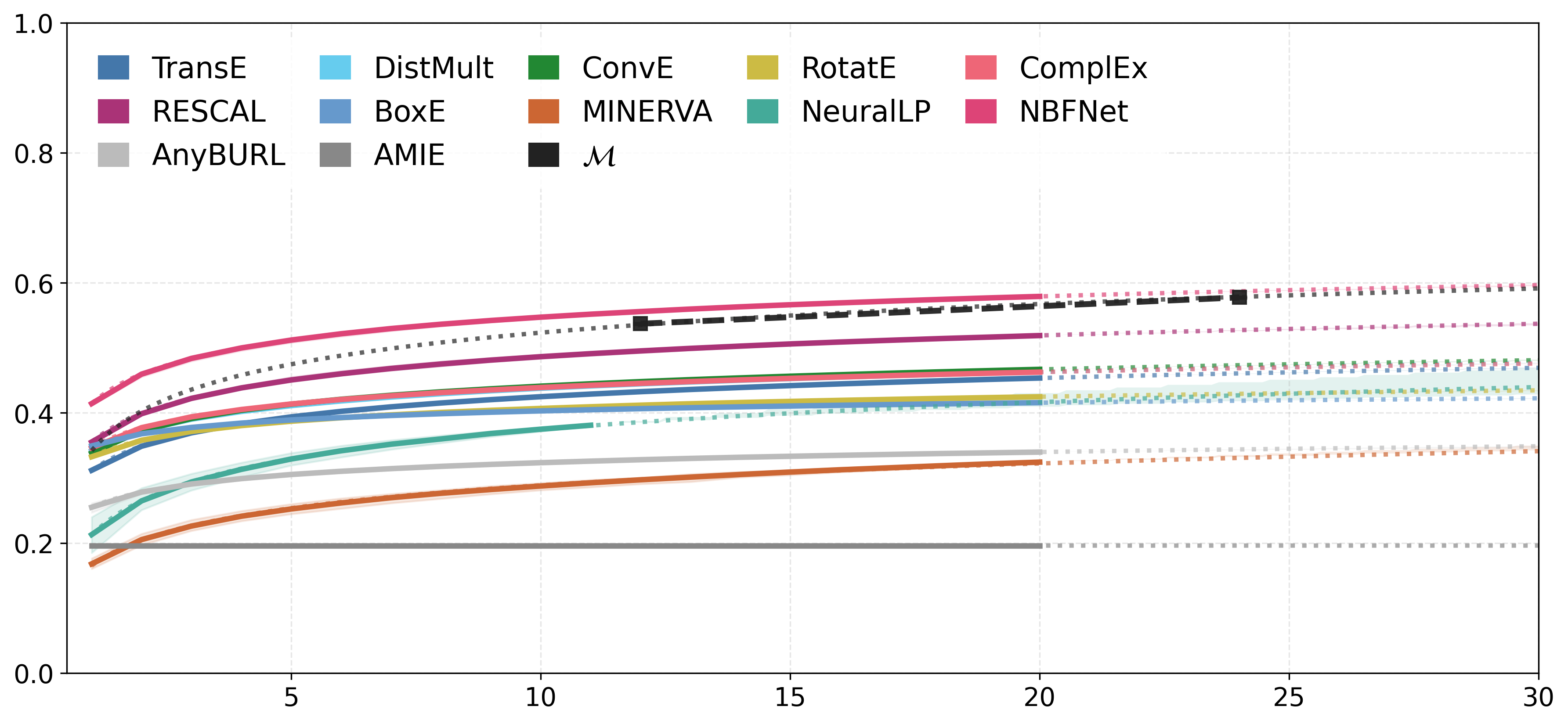}
    \caption{Evolution of oracle MRR as a function of the number of instances $k$ on FB15k-237.}

    \label{fig:dynamic_mrr_saturation}
\end{figure}

To address this, we study the evolution of $\mathrm{MRR}^{k}_{\mathcal{O}_{G}}$ as a function of the number of instances $k$. Figure~\ref{fig:dynamic_mrr_saturation} shows the observed evolution of $\mathrm{MRR}^{k}_{\mathcal{O}_{M}}$ on FB15k-237 for each model $M \in \mathcal{M}$ (solid lines) up to our empirical limit of $k \leq 20$. We additionally report $\mathrm{MRR}^{12}_{\mathcal{O}_{\mathcal{M}}}$ and $\mathrm{MRR}^{24}_{\mathcal{O}_{\mathcal{M}}}$, corresponding to cross-model groups with one and two instances per model, respectively. Similar trends are observed on other datasets (see GitHub\textsuperscript{\ref{footnote:github}}). Because multiple instance orderings are possible, we sample $100$ different permutations and report the mean trajectory, along with the best and worst cases. 
Results indicate that oracle performance depends primarily on the number of instances rather than their ordering, reinforcing the view that diversity is uniformly distributed across runs rather than concentrated in a few specific instances.

\begin{figure}
    \centering
    \includegraphics[width=1\linewidth]{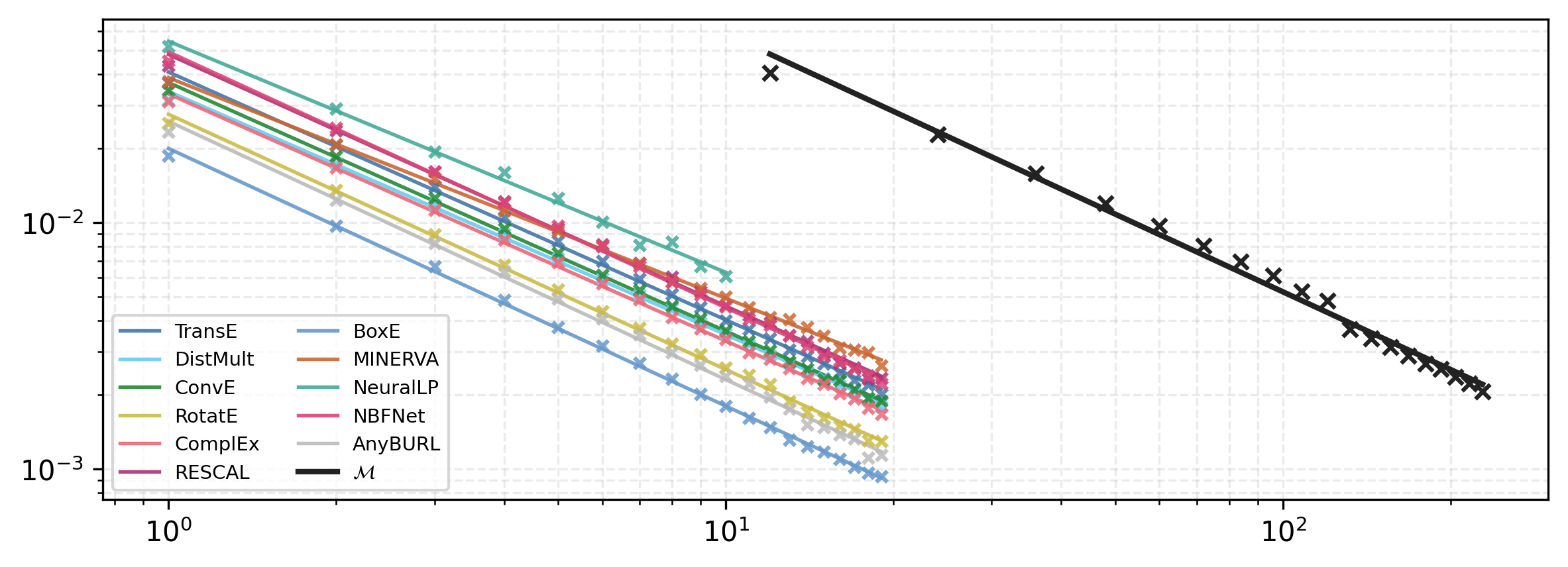}
    \caption{Power-law fits of $\Delta \mathrm{MRR}(k)$ (y-axis) w.r.t. instance number (x-axis) in log-log scale across models on FB15k-237.}
    \label{fig:loglog_fit}
\end{figure}

Overall, the figure highlights a clear diminishing-return effect: the marginal gain of each additional instance decreases rapidly as $k$ increases. To characterize this trend beyond the observed $k \leq 20$, we model the marginal performance gain using a power law:
\[
\Delta \mathrm{MRR}(k)
=
\mathrm{MRR}^{k+1}_{\mathcal{O}_{G}}
-
\mathrm{MRR}^{k}_{\mathcal{O}_{G}}
=
A k^{-m}.
\]
Taking the logarithm of both sides gives
$
\log \Delta \mathrm{MRR}(k)
=
\log A - m \log k$.
We therefore estimate $\log A$ and $m$ using ordinary least-squares regression on the log-transformed observations. The resulting fits obtain $R^2 > 0.99$ across almost all considered models\footnote{The point at which NeuralLP ceases to contribute can be observed in $\Delta \mathrm{MRR}_{G_{\mathcal{M}}}$.}, as illustrated in Figure~\ref{fig:loglog_fit}.

For a given number of instances $k$, the oracle performance can then be reconstructed by summing the marginal gains:
\[
\mathrm{MRR}^{k}_{\mathcal{O}_{G}}
=
\mathrm{MRR}^{1}_{\mathcal{O}_{G}}
+
\sum_{i=1}^{k-1} \Delta \mathrm{MRR}(i)
\approx
\mathrm{MRR}^{1}_{\mathcal{O}_{G}}
+
A\sum_{i=1}^{k-1}i^{-m}.
\]
This provides a simple extrapolation of oracle performance beyond the observed range.

Crucially, we fit $(A,m)$ separately for each of the $100$ sampled permutations rather than on their mean trajectory. This turns the extrapolation into a distribution of possible performance trajectories, from which we report the median and $95\%$ interval. The resulting extrapolations are shown as dashed curves in Figure~\ref{fig:dynamic_mrr_saturation}.

The extrapolations suggest that an unrealistically large number of instances would be required to approach a performance of $1$. To further highlight this phenomenon, we use this extrapolation in Table~\ref{tab:f_10000_mrr_scaling_law} to estimate the oracle performance under an extreme scenario in which the number of instances reaches $k=10{,}000$. 
The extrapolated results indicate that, for most groups, including the cross-model group $G_{\mathcal{M}}$, even such an extreme number of instances would not bring oracle performance close to $1$. A few exceptions are observed, notably for MINERVA and NeuralLP on FB15k-237 and for NeuralLP and NBFNet on CoDEx-S. In these cases, the observed gains vary significantly across instances, resulting in a high sensitivity of the extrapolation to the permutations used for fitting the scaling law and consequently in wide prediction intervals. Some extrapolations therefore reach $1$, which should not be interpreted as evidence that perfect performance would actually be achieved. ConvE on CoDEx-S constitutes the notable exception, with high performance being extrapolated.

The extrapolation nevertheless only provides an indication of how the observed scaling trend may evolve beyond the number of instances that we actually evaluate. The underlying phenomenon is already directly observable with the $240$ instances considered in the largest group, $G_{\mathcal{M}}$: the oracle fails to resolve a non-negligible subset of queries. Thus, even when exploiting substantial stochastic and architectural diversity, some queries remain unresolved by all of the considered models.

These findings highlight the third gap identified in this paper: the knowledge captured collectively by the considered models remains incomplete. The unresolved queries may reflect limitations in the expressivity of the considered approaches or deficiencies in the benchmarks, which may contain information that is fundamentally unrecoverable by any statistical model.

In Appendix~\ref{app:hard_queries}, we investigate what distinguishes these hard-to-predict queries from the others, particularly those that are commonly easy to solve.

\begin{table}
\centering
\caption{Median estimated value of $\mathrm{MRR}^{10{,}000}_{\mathcal{O}_{G}}$ using the marginal-gain scaling over $100$ permutations, with the $95\%$ interval in brackets. $\dagger$ marks groups for which some permutations extrapolate above $1$.}

\small
\setlength{\tabcolsep}{3pt}
\begin{tabular}{lccc}
\toprule
Model & FB15k-237 & WN18RR & CoDEx-S \\
\midrule
TransE & 0.70 {\tiny [0.67, 0.72]} & 0.31 {\tiny [0.30, 0.33]} & 0.51 {\tiny [0.49, 0.58]} \\
DistMult & 0.69 {\tiny [0.67, 0.72]} & 0.60 {\tiny [0.58, 0.65]} & 0.77 {\tiny [0.71, 0.90]} \\
ConvE & 0.68 {\tiny [0.66, 0.72]} & 0.72 {\tiny [0.68, 0.77]} & $0.92^{\dagger}$ {\tiny [0.85, 0.99]} \\
RotatE & 0.56 {\tiny [0.54, 0.59]} & 0.63 {\tiny [0.60, 0.66]} & 0.70 {\tiny [0.64, 0.79]} \\
ComplEx & 0.66 {\tiny [0.65, 0.69]} & 0.77 {\tiny [0.71, 0.87]} & 0.79 {\tiny [0.76, 0.84]} \\
RESCAL & 0.77 {\tiny [0.72, 0.85]} & 0.83 {\tiny [0.77, 0.94]} & 0.71 {\tiny [0.62, 0.95]} \\
BoxE & 0.51 {\tiny [0.49, 0.53]} & 0.61 {\tiny [0.59, 0.67]} & 0.57 {\tiny [0.54, 0.73]} \\
MINERVA & $0.76^{\dagger}$ {\tiny [0.62, 1.10]} & 0.54 {\tiny [0.52, 0.63]} & 0.71 {\tiny [0.59, 0.91]} \\
NeuralLP & $0.90^{\dagger}$ {\tiny [0.62, 1.73]} & 0.56 {\tiny [0.55, 0.58]} & $0.84^{\dagger}$ {\tiny [0.57, 2.66]} \\
NBFNet & 0.82 {\tiny [0.78, 0.86]} & 0.75 {\tiny [0.73, 0.79]} & $0.78^{\dagger}$ {\tiny [0.66, 1.49]} \\
AnyBURL & 0.45 {\tiny [0.42, 0.51]} & 0.50 {\tiny [0.49, 0.51]} & 0.31 {\tiny [0.31, 0.32]} \\
\midrule
$\mathcal{M}$ & 0.84 {\tiny [0.82, 0.85]} & 0.81 {\tiny [0.79, 0.84]} & 0.87 {\tiny [0.83, 0.91]} \\
\bottomrule
\end{tabular}
\label{tab:f_10000_mrr_scaling_law}
\end{table}

\section{Discussion}
\label{sec:discussion}

Our proposed oracle-framework provides upper bounds that could be reached by ensemble of instances.
While we do not evaluate existing ensemble methods on our instances, the improvements reported in prior work~\cite{GregucciN0S23,YueZYLWZL023} appear smaller than those suggested by oracles. This observation indicates that additional gains may still be attainable through more effective selection or combination strategies.
In this view, oracles provide useful reference points for future ensemble learning benchmarks, as they correspond to an ideal routing mechanism. 
In principle, combining predictions could even surpass this bound, as certain aggregation mechanisms (e.g., voting or score weighting and aggregation between instances) may recover correct rankings that are not produced by any individual instance for a given query.

In our experimental setting, instance diversity is induced through different random seeds or model architectures, while keeping hyperparameter configurations fixed. 
Traditionally, hyperparameter tuning is designed to optimize the performance of individual instances, whereas our analysis focuses on the collective performance of groups. 
Hyperparameter selection could thus be alternatively designed to promote diversity at the group level: either by identifying a single configuration that promotes diversity across instances, or by intentionally varying configurations to generate complementary behaviors. Exploring these alternatives could provide an additional source of diversity for improving ensemble performance.

Regarding our observation that all models tend to fail on the same subset of queries, one may wonder whether this is a consequence of training all instances on the same data distribution. In such a setting, models may learn a shared core signal, with diversity emerging only at the margins. 
Alternative strategies could thus be envisioned to further enhance ensemble effectiveness, such as encouraging specialization by training instances on different regions of the knowledge graph or introducing coordination mechanisms that promote complementary behaviors.
Alternatively, this could also constitute a potential limitation of current benchmarks, which may include queries that are inherently unanswerable given the available training signal. In such cases, achieving perfect performance would be impossible, regardless of the modeling approach.

Finally, approaching oracle performance requires identifying \emph{a priori} which instance is most likely to produce the best prediction for a given query. The next step therefore lies in identifying performance and complementarity signals to guide instance selection or weighting in ensemble approaches.

\section{Conclusion}
\label{sec:conclusion}

In this paper, we propose an oracle framework as an idealized setting that provides an upper-bound on ensemble learning performance for link prediction. 
This framework reveals three performance gaps.
First, ensembling multiple instances of the same model, trained with different random seeds, yields substantial performance gains, showcasing that isolated instances are not representative of the attainable performance of a model.
Second, aggregating instances across different model architectures matches or outperforms the best single-model oracle, illustrating that no approach provides alone the diversity of combining various architectures, regardless of their individual performance.
Third, even with increasingly large ensembles, performance consistently converges below perfect accuracy, indicating limitations in knowledge capture by current approaches or in learnable knowledge in current benchmarks.
These gaps underscore the value of ensemble learning for link prediction in knowledge graphs, while opening several directions for future work: designing single-model or cross-model aggregation strategies to attain oracle performance, increasing diversity in model instances to boost performance, and identifying unlearnable queries in benchmark datasets. 

\newpage

\bibliographystyle{splncs04}
\bibliography{mybibliography}

\newpage

\appendix

\section{Characterizing Difficult Queries}
\label{app:hard_queries}

To understand why some queries remain unsolved by all instances, we analyze whether difficult queries exhibit
different structural properties from queries that are consistently solved.

\paragraph{Query difficulties.}
We classify queries by how consistently instances in $G_{\mathcal{M}}$ recover
the correct answer at rank~1. A query is \emph{Easy} if every instance ranks
the correct answer first, \emph{Hard} if none does, and \emph{Medium} otherwise.
To prevent low-performing models from shrinking the global Easy set (e.g., TransE on WN18RR), for each dataset, we compute the Easy set of each model separately and exclude those whose Easy ratio is below $15\%$ .

\paragraph{Query features.}
We characterize each query using two families of features computed from
the training graph: \emph{graph connectivity} and \emph{relation
structure}.
For \emph{graph connectivity}, consider a query $(h,r,?)$ whose ground-truth
triple is $(h,r,t)$. We measure how frequently the target entity $t$
has been observed as an object of relation $r$ in the training graph
(\emph{target occurrence, per relation}), and how frequently the source
entity $h$ has been observed as a subject of relation $r$
(\emph{source occurrence, per relation}). The latter is equivalent to
the number of valid answers for the query in the training graph as, for
$(h,r,?)$, it counts the distinct entities $t'$ such that $(h,r,t')$
occurs in the training set.
For \emph{relation structure}, we consider simple relation patterns
(see, e.g., Liu et al.~\cite{Horrocks}). In particular, a query
$(h,r,?)$ whose ground-truth triple is $(h,r,t)$ is considered
\emph{symmetric} if the relation satisfies the symmetry pattern
$(x,r,y) \Rightarrow (y,r,x)$ with confidence greater than $80\%$ in
the training graph, and the corresponding mirror triple $(t,r,h)$ is
observed in the training set. We also characterize each relation by
its cardinality type (\emph{N-1}, or \emph{N-N}).
Note that we considered additional features, including global entity
occurrence, 2- and 3-hop paths, relation composition, 1-1 and 1-N cardinality types.
We focus here on the features with the clearest signal in our analysis, but provide the full set of features on GitHub\textsuperscript{\ref{footnote:github}}.
For each feature, we report its value separately for Easy, Medium, and
Hard queries, as well as over all queries (\emph{Total}). Numerical
features are summarized by their median, while relation patterns are
reported as the percentage of queries exhibiting the corresponding
property.

To quantify how strongly a feature distinguishes Easy from Hard
queries, we compute the area under the receiver operating
characteristic curve (AUC). For a feature $f$, the AUC
can be interpreted~\footnote{\url{https://en.wikipedia.org/wiki/Mann-Whitney_U_test}} as the probability that a randomly selected Easy
query receives a higher feature value than a randomly selected Hard
query, with ties counted as half:
\[
\mathrm{AUC}(f)
=
\Pr\left[f(q_E)>f(q_H)\right]
+
\frac{1}{2}
\Pr\left[f(q_E)=f(q_H)\right],
\]
where $q_E$ and $q_H$ are independently sampled Easy and Hard queries. An AUC of $1$ therefore means that the feature perfectly separates the
two groups, with larger values systematically associated with Easy
queries. Conversely, an AUC of $0$ corresponds to perfect separation in
the opposite direction, with larger values associated with Hard queries.
An AUC of $0.5$ indicates that the feature provides no discrimination
between the two groups. The AUC furthest from $0.5$ for each dataset is shown in bold in Table~\ref{tab:hard_query_profile}.

\begin{table}[t]
\centering
\caption{Characteristics of Easy, Medium, and Hard queries.}

\setlength{\tabcolsep}{4pt}
\resizebox{0.9\linewidth}{!}{\begin{tabular}{llrrrrc}
\toprule
Dataset & Statistic & Easy & Medium & Hard & Total & AUC \\
\midrule
\multirow{5}{*}{FB15k-237}
& \% of queries & 10\% & 49\% & 41\% & 100\% & -- \\
& Target occ.\ (rel.) & 966 & 22 & 5 & 12 & \textbf{0.952} \\
& Source occ.\ (rel.) & 1 & 11 & 27 & 12 & 0.135 \\
& Symmetry & 0.0\% & 0.0\% & 0.0\% & 0.0\% & 0.500 \\
& N-1 & 57.5\% & 11.9\% & 6.2\% & 14.1\% & 0.757 \\
& N-N & 41.2\% & 76.0\% & 71.8\% & 70.8\% & 0.347 \\
\midrule
\multirow{5}{*}{WN18RR}
& \% of queries & 21\% & 47\% & 32\% & 100\% & -- \\
& Target occ.\ (rel.) & 2 & 2 & 1 & 2 & 0.589 \\
& Source occ.\ (rel.) & 2 & 2 & 1 & 2 & 0.501 \\
& Symmetry & 97.8\% & 33.4\% & 0.0\% & 36.0\% & \textbf{0.989} \\
& N-1 & 1.2\% & 33.1\% & 43.7\% & 30.0\% & 0.287 \\
& N-N & 97.0\% & 36.3\% & 4.6\% & 38.6\% & 0.962 \\
\midrule
\multirow{5}{*}{CoDEx-S}
& \% of queries & 6\% & 56\% & 37\% & 100\% & -- \\
& Target occ.\ (rel.) & 516 & 38 & 8 & 22 & \textbf{0.973} \\
& Source occ.\ (rel.) & 3 & 18 & 53 & 22 & 0.138 \\
& Symmetry & 0.0\% & 24.5\% & 0.3\% & 13.9\% & 0.499 \\
& N-1 & 42.8\% & 12.6\% & 3.3\% & 11.1\% & 0.697 \\
& N-N & 57.2\% & 79.7\% & 77.6\% & 77.5\% & 0.398 \\
\bottomrule
\end{tabular}}
\label{tab:hard_query_profile}

\end{table}

\paragraph{Analysis.}
The factors associated with query difficulty differ substantially across datasets. On FB15k-237 and CoDEx-S, difficulty is primarily driven by entity connectivity. Target occurrence is the most discriminative feature, with median values dropping from 966 to 5 on FB15k-237 (AUC~$0.95$) and from 516 to 8 on CoDEx-S (AUC~$0.97$) between Easy and Hard queries. Thus, queries are easier when the ground-truth target has frequently been observed as an object of the query relation in the training graph. Conversely, the number of valid answers increases from 1 to 27 and from 3 to 53, respectively (AUC~$0.14$ in both cases), indicating that queries with more possible answers are harder to solve. Relation patterns are substantially less discriminative on these datasets, with maximum AUCs of $0.76$ and $0.70$, respectively.

WN18RR exhibits the opposite pattern: connectivity features are weak (maximum AUC~$0.62$), while relation structure is highly discriminative. Symmetry reaches an AUC of $0.99$, with $97.8\%$ of Easy queries having a training-set witness for the mirror triple $(t,r,h)$, compared with none of the Hard queries. N-N cardinality also reaches an AUC of $0.96$.

Overall, query difficulty depends strongly on the structure of the underlying KG, with entity connectivity dominating on FB15k-237 and CoDEx-S, and relation structure dominating on WN18RR.

\section{UpSet plot}
\label{app:upset}

\begin{figure*}[t]
    \centering
    \begin{subfigure}{0.48\linewidth}
        \centering
        \includegraphics[width=\linewidth]{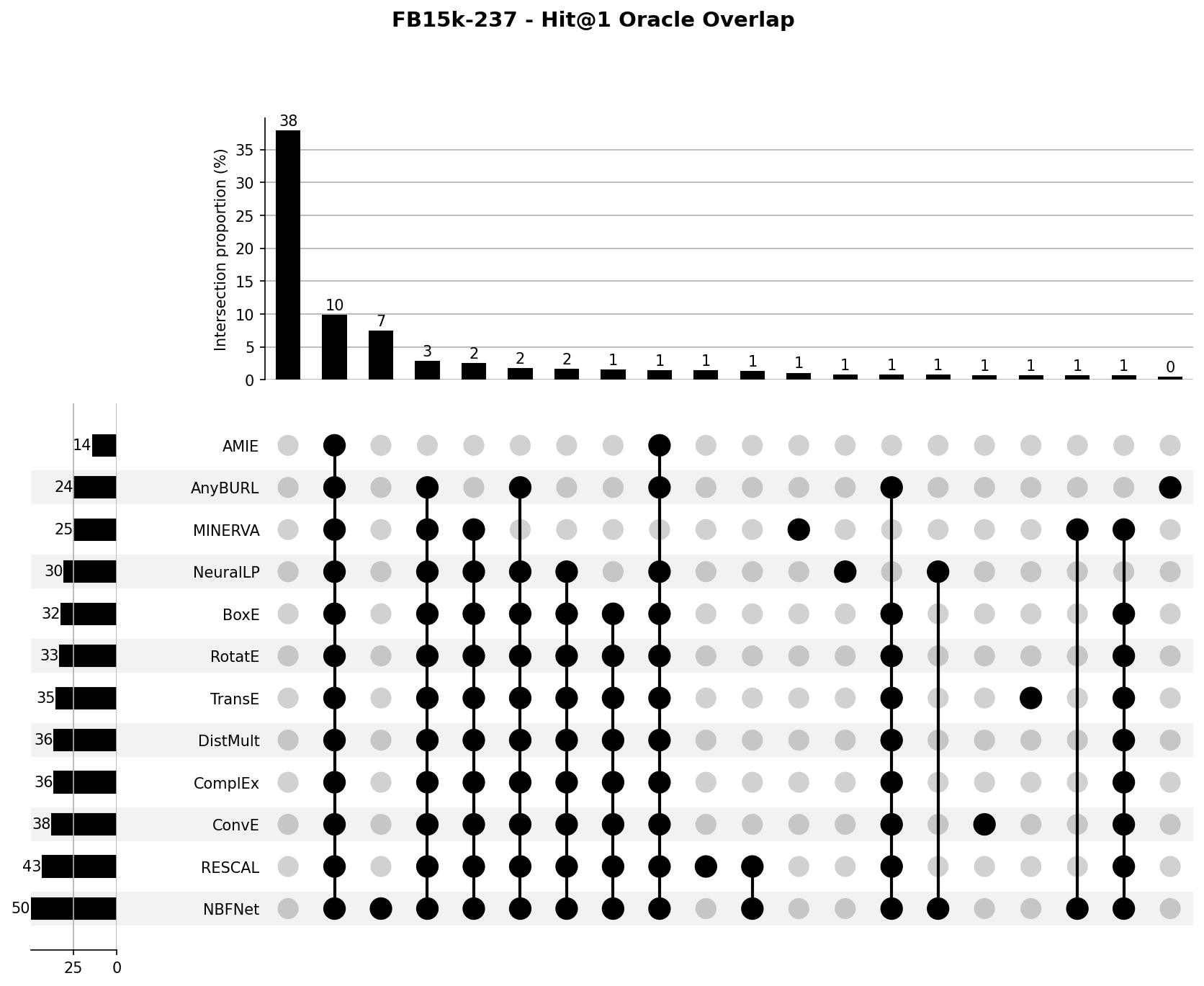}
        \caption{FB15k-237 (Hit@1)}
        \label{fig:fb_upset_hit1}
    \end{subfigure}
    \hfill
    \begin{subfigure}{0.48\linewidth}
        \centering
        \includegraphics[width=\linewidth]{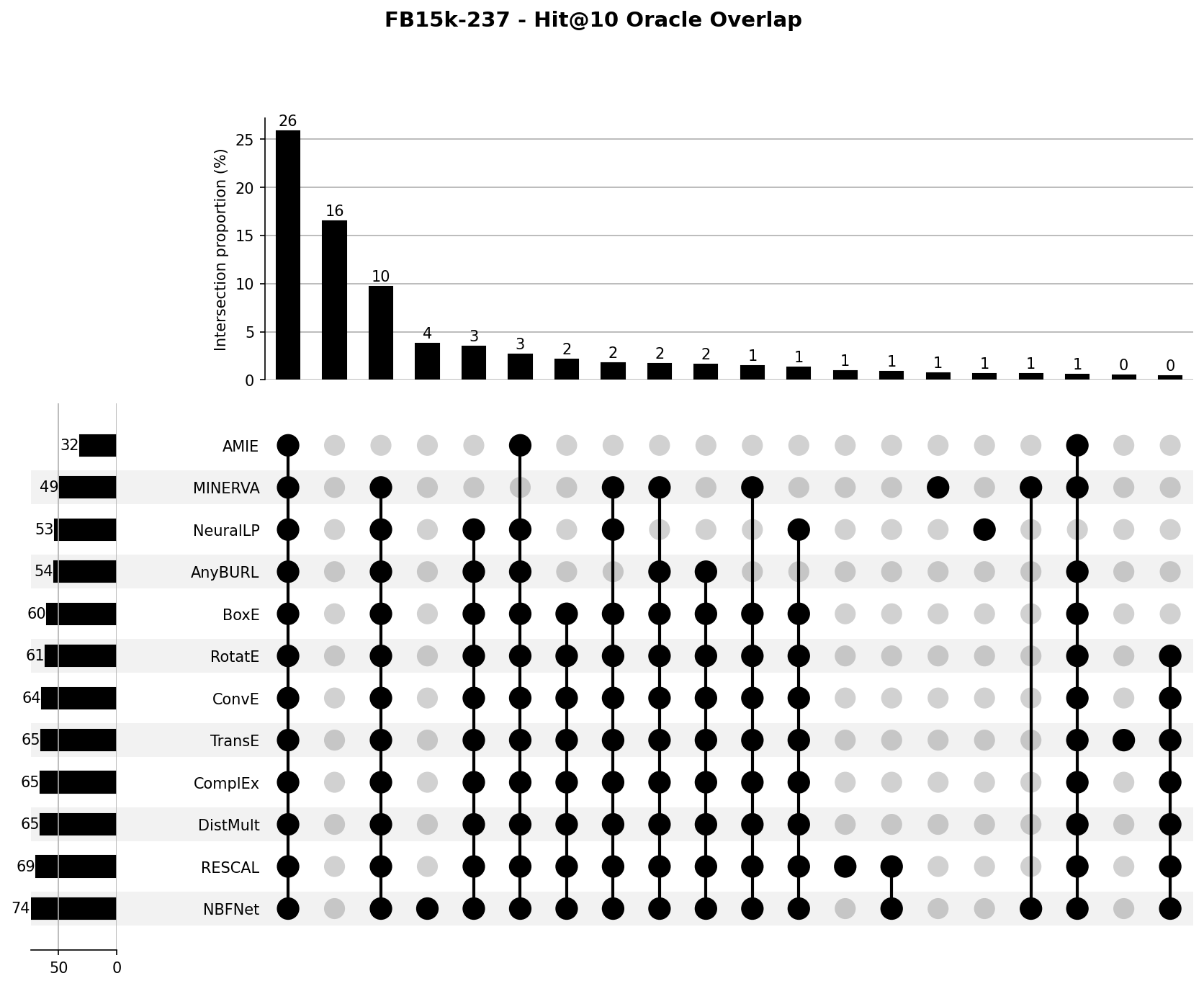}
        \caption{FB15k-237 (Hit@10)}
        \label{fig:fb_upset_hit10}
    \end{subfigure}

    \vspace{0.5em}

    \begin{subfigure}{0.48\linewidth}
        \centering
        \includegraphics[width=\linewidth]{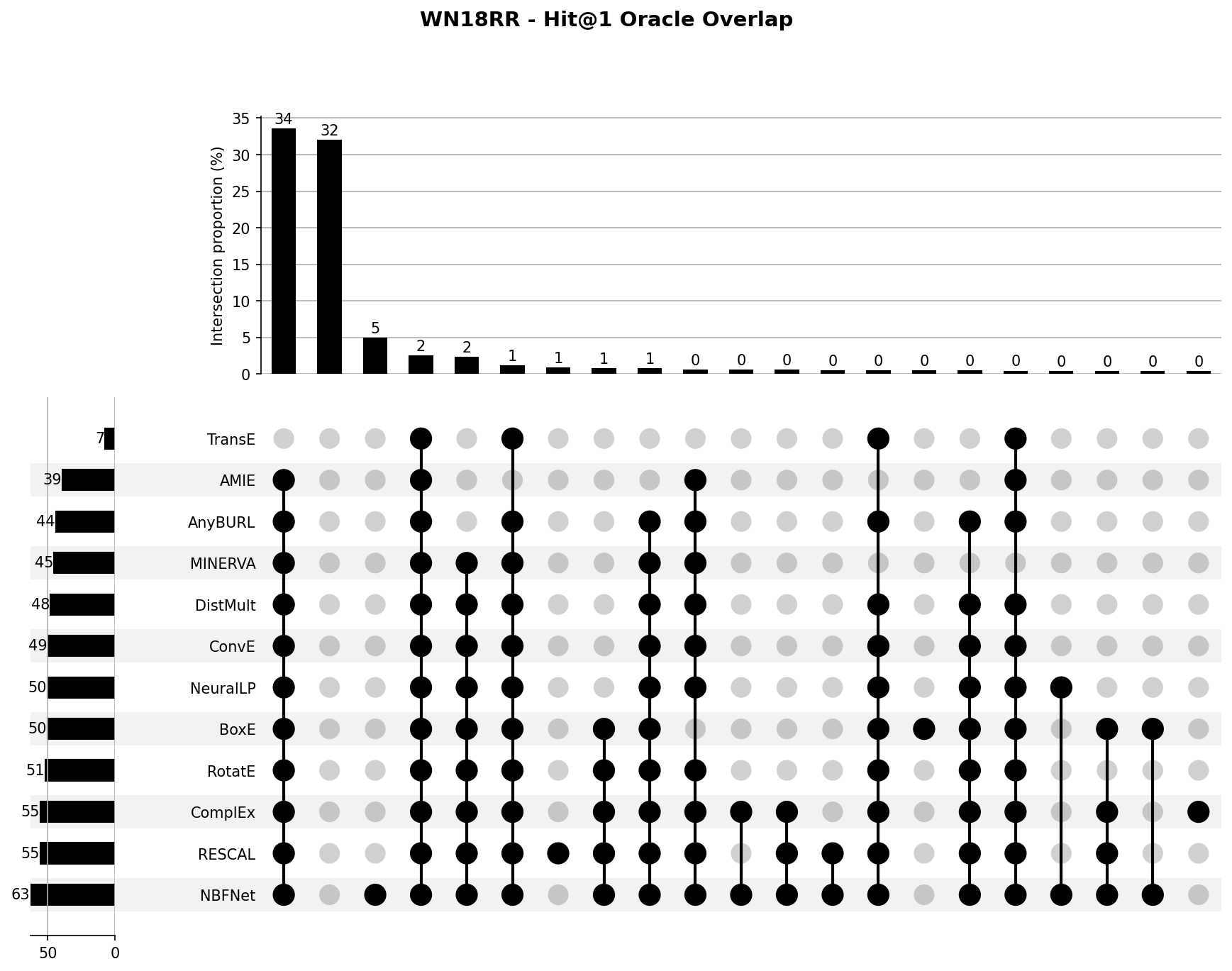}
        \caption{WN18RR (Hit@1)}
        \label{fig:wn_upset_hit1}
    \end{subfigure}
    \hfill
    \begin{subfigure}{0.48\linewidth}
        \centering
        \includegraphics[width=\linewidth]{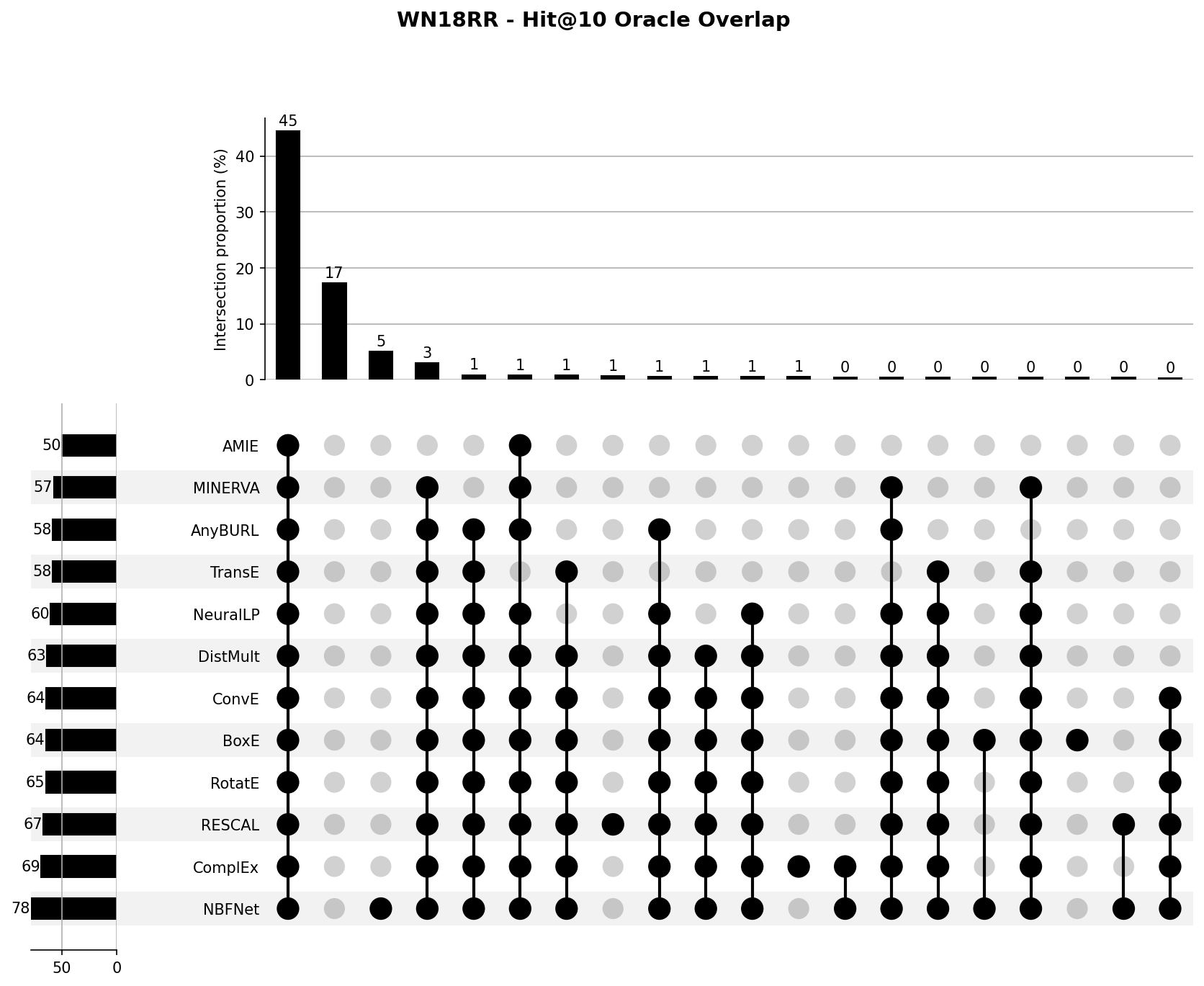}
        \caption{WN18RR (Hit@10)}
        \label{fig:wn_upset_hit10}
    \end{subfigure}

    \caption{UpSet plots showing the distribution of correctly answered queries across the 20 most frequent combinations of models for FB15k-237 and WN18RR. Each bar represents the proportion of queries solved by a specific subset of models, enabling the analysis of overlap and model complementarity at different evaluation thresholds (Hit@1 and Hit@10).}
    \label{fig:upsetplotFB}
\end{figure*}

The UpSet plots in Fig.~\ref{fig:upsetplotFB} provide a complementary perspective to Fig.~\ref{fig:count_hit_model}, offering a more fine-grained analysis of model behavior. Specifically, they depict the proportion of queries successfully answered (i.e., $\text{Hit@1}=1$ or $\text{Hit@10}=1$) for the 20 most represented combinations of models. In other words, each bar corresponds to the set of queries that are jointly solved by a given subset of models and by no others, thereby revealing patterns of overlap, exclusivity, and complementarity among models. Results for CoDEx-S are available in our GitHub repository under \texttt{oracles\_outputs/histogram}.

For instance, considering Hit@1 on FB15k-237, we observe that $38\%$ of the queries are not correctly answered by any model, while $10\%$ are successfully solved by all models. Furthermore, $7\%$ of the queries are uniquely solved by NBFNet, highlighting its distinctive contribution. Interestingly, adding any of RESCAL, MINERVA, NeuralLP, or even TransE to the other models increases the overall number of solved queries by more than $1\%$, despite their overall lower performance. This suggests that even weaker models can capture complementary signals that are not exploited by stronger ones.

\section{Results for Hit@1}
\label{app:reshit}
\begin{table}
\centering
\caption[]{Averaged and Oracle Hit@1 across datasets and models}
\small
\setlength{\tabcolsep}{3pt}
\begin{tabular}{cc lccc}
\toprule
 & & Model & WN18RR & FB15k-237 & codex-s \\
\midrule

\multirow{20}{*}{\rotatebox{90}{\textbf{Neural}}} 
& \multirow{14}{*}{\rotatebox{90}{KGE}}
& $\overline{\mathrm{MRR}}_{G_{\text{TransE}}}$ & $.06 \pm .00$ & $.22 \pm .00$ & $.22 \pm .00$ \\
& & $\mathrm{MRR}_{\mathcal{O}_{\text{TransE}}}$ & $.07$ & $.35$ & $.27$ \\
& & $\overline{\mathrm{MRR}}_{G_{\text{DistMult}}}$ & $.44 \pm .00$ & $.25 \pm .00$ & $.30 \pm .01$ \\
& & $\mathrm{MRR}_{\mathcal{O}_{\text{DistMult}}}$ & $.48$ & $.36$ & $.46$ \\
& & $\overline{\mathrm{MRR}}_{G_{\text{ConvE}}}$ & $.44 \pm .00$ & $.25 \pm .00$ & $.34 \pm .01$ \\
& & $\mathrm{MRR}_{\mathcal{O}_{\text{ConvE}}}$ & $.49$ & $.38$ & $.55$ \\
& & $\overline{\mathrm{MRR}}_{G_{\text{RotatE}}}$ & $.47 \pm .00$ & $.24 \pm .00$ & $.30 \pm .00$ \\
& & $\mathrm{MRR}_{\mathcal{O}_{\text{RotatE}}}$ & $.51$ & $.33$ & $.43$ \\
& & $\overline{\mathrm{MRR}}_{G_{\text{ComplEx}}}$ & $.47 \pm .00$ & $.25 \pm .00$ & $.37 \pm .00$ \\
& & $\mathrm{MRR}_{\mathcal{O}_{\text{ComplEx}}}$ & $.55$ & $.36$ & $.51$ \\
& & $\overline{\mathrm{MRR}}_{G_{\text{RESCAL}}}$ & $.47 \pm .00$ & $.26 \pm .00$ & $.29 \pm .00$ \\
& & $\mathrm{MRR}_{\mathcal{O}_{\text{RESCAL}}}$ & $.55$ & $.43$ & $.44$ \\
& & $\overline{\mathrm{MRR}}_{G_{\text{BoxE}}}$ & $.43 \pm .00$ & $.25 \pm .00$ & $.31 \pm .01$ \\
& & $\mathrm{MRR}_{\mathcal{O}_{\text{BoxE}}}$ & $.50$ & $.32$ & $.39$ \\
\cmidrule{2-6}

& & $\overline{\mathrm{MRR}}_{G_{\text{MINERVA}}}$ & $.39 \pm .01$ & $.11 \pm .00$ & $.19 \pm .02$ \\
& & $\mathrm{MRR}_{\mathcal{O}_{\text{MINERVA}}}$ & $.45$ & $.25$ & $.35$ \\
& & $\overline{\mathrm{MRR}}_{G_{\text{NeuralLP}}}$ & $.31 \pm .05$ & $.16 \pm .01$ & $.19 \pm .01$ \\
& & $\mathrm{MRR}_{\mathcal{O}_{\text{NeuralLP}}}$ & $.50$ & $.30$ & $.35$ \\
& & $\overline{\mathrm{MRR}}_{G_{\text{NBFNet}}}$ & $.53 \pm .00$ & $.32 \pm .00$ & $.26 \pm .01$ \\
& & $\mathrm{MRR}_{\mathcal{O}_{\text{NBFNet}}}$ & $.63$ & $.50$ & $.42$ \\
\midrule

\multirow{4}{*}{\rotatebox{90}{\textbf{Symbolic}}}
& & $\overline{\mathrm{MRR}}_{G_{\text{AnyBURL}}}$ & $.43 \pm .00$ & $.18 \pm .00$ & $.20 \pm .00$ \\
& & $\mathrm{MRR}_{\mathcal{O}_{\text{AnyBURL}}}$ & $.44$ & $.24$ & $.20$ \\
& & $\overline{\mathrm{MRR}}_{G_{\text{AMIE}}}$ & $.39 \pm .00$ & $.14 \pm .00$ & $.15 \pm .00$ \\
& & $\mathrm{MRR}_{\mathcal{O}_{\text{AMIE}}}$ & $.39$ & $.14$ & $.15$ \\
\midrule

\multirow{9}{*}{\rotatebox{90}{\textbf{Cross-model}}}
& & $\mathrm{MRR}^{21}_{\mathcal{O}_{KGE}}$ & $.57 \pm .00$ & $.42 \pm .00$ & $.55 \pm .00$ \\ \addlinespace
& & $\mathrm{MRR}^{20}_{\mathcal{O}_{Neural}}$ & $.62 \pm .00$ & $.49 \pm .00$ & $.58 \pm .00$ \\ \addlinespace
& & $\mathrm{MRR}^{20}_{\mathcal{O}_{Symbolique}}$ & $.44 \pm .00$ & $.24 \pm .00$ & $.21 \pm .00$ \\ \addlinespace
& & $\mathrm{MRR}^{24}_{\mathcal{O}_{\mathcal{M}}}$ & $.62 \pm .00$ & $.50 \pm .00$ & $.58 \pm .00$ \\ \addlinespace
& & $\mathrm{MRR}^{20, \text{greedy}}_{\mathcal{O}_{\mathcal{M}}}$ & $.64 \pm .00$ & $.52 \pm .00$ & $.59 \pm .00$ \\ \addlinespace
& & $\mathrm{MRR}^{240}_{\mathcal{O}_{\mathcal{M}}}$ & $.68$ & $.62$ & $.69$ \\

\bottomrule
\end{tabular}
\label{tab:rq1_hits_at_1}
\end{table}

\section*{GenAI Usage Disclosure}
During the preparation of this paper, ChatGPT was used as a writing assistant to improve language. In addition, Windsurf Editor, powered by Claude Sonnet 4.6, was used as a coding assistant.
All scientific contributions, analyses, and experimental results presented in this paper are the authors' own.

% \bibliographystyle{splncs04}
% \bibliography{mybibliography}

\end{document}